# Low-Rank Dynamics-Effective Latent Carriers for Counterfactual Rollout in Learned World Models

Yang Liu*, Yuming Chen*
College of Intelligent Robotics and Advanced Manufacturing, Fudan University
Shanghai, China
{ly, yumingchen}@fudan.edu.cn

## Abstract

We ask whether a small, directly addressable hidden-state intervention can place a learned world model on an intended counterfactual future and then let the model's own dynamics carry that future forward. In a controlled two-object collision environment, we study a 192-dimensional recurrent model trained on factual and locally edited counterfactual trajectories. Candidate carriers are learned from training-only counterfactual-minus-factual hidden differences, and an affine map predicts carrier coordinates from the factual state and requested edit without access to the native counterfactual hidden state at test time. For bounded single-component velocity edits, rank 4 is the smallest tested rank on the preregistered grid that satisfies the development criteria. A one-shot rank-4 patch launches a 12-transition autonomous rollout without future observations, teacher forcing, repeated hidden-state correction, or physical-state clamping. The frozen procedure satisfies the preregistered 2-of-3 fresh-checkpoint replication rule and remains reusable at nearby anchors. The same Single-derived carrier and Single-only affine map also support bounded same-object two-component requests. Across the matched training regimes, broader counterfactual support was associated mainly with better Joint rollout accuracy and more additive Joint hidden responses. Composition-related structure is enriched in the rank-4 subspace but is not confined to it, and local recurrent diagnostics show strong one-step coupling from the carrier to the rest of the hidden state. A position-edit stress test fails the required specificity controls. Together, these results support a compact dynamics-effective intervention-entry interface, not a closed four-dimensional state or an intrinsic state dimension.

## 1 Introduction

World models have progressed from compact learned simulators (Ha and Schmidhuber 2018) and latent planning (Hafner et al. 2019) to imagination-based control (Hafner et al. 2023) , generative interactive environments (Bruce et al. 2024) , and action-conditioned video models for planning (Assran et al. 2025). This advance makes latent state itself a scientific object: not only whether a model predicts, but what its internal state retains and whether that information can guide future behavior.

The key question is whether the latent state can support the intended counterfactual behavior, not whether its coordinates have an obvious physical interpretation. DeepMDP formalized latent compression that preserves transition and reward structure relevant to control (Gelada et al. 2019). Recent work argues that latent states should be judged by sufficiency for a specified function, such as prediction, planning, memory, grounding, or counterfactual reasoning (K.W. Kim 2026), and that world models should preserve distinctions needed to answer intervention queries (Thorpe et al. 2026). Counterfactual behavior therefore provides a stricter test than decodability alone.

---

* Equal contribution

Source codes and data can be found at https://github.com/lysea8282/dynamic-effective-latent-carriers

Physical dynamics sharpen the problem. Physion and related benchmarks show that learned models can make useful physical predictions (Bear et al. 2021), but behavioral success does not reveal how task-relevant information is organized internally. A decodable physical quantity need not be used by the transition dynamics, while functionally relevant state may be distributed across hidden coordinates.

A useful analogy is coarse-graining: complex systems can admit coarser predictive descriptions, as in rigorous micro-to-macro limits such as hard-sphere dynamics to fluid equations (Deng, Hani, and Ma 2025). The analogy is inspiring but has limits. Neural latent coordinates are learned and non-identifiable, so a useful intervention interface need not align with named physical variables or a fixed set of human-interpretable axes.

Recent interpretability work makes this distinction concrete. Distributed Alignment Search allows causal variables to align with distributed neural representations (Geiger et al. 2024) , while causal-probing work emphasizes completeness and selectivity (Canby et al. 2025). In world and video models, physical information has been probed in distributed representations (Joseph et al. 2026), probe-derived hidden-state shifts can alter predictions (Zhang 2026), and Concept Activation Vectors can steer physical plausibility judgments (Alam 2026). These results establish that physical information can be readable and steerable, but they do not by themselves show that an edited hidden state can autonomously generate the intended future.

Our central question then asks: can a limited hidden-state intervention place a recurrent world model on an intended counterfactual trajectory and then let the model's own dynamics continue it? We require a one-shot edit to produce a sustained autonomous rollout that passes registered specificity and control tests, rather than merely changing an immediate readout or one-step prediction.

We study a controlled 192-dimensional recurrent world model of two-object motion in a deterministic two-dimensional collision environment. The primary model family is trained on factual trajectories together with locally edited counterfactual trajectories. We call bounded edits to one velocity component “Single” and same-object edits to two velocity components “Joint.” After verifying that the model can natively roll out the edited state, we fit checkpoint-specific candidate carriers from training-only counterfactual-minus-factual hidden differences and an affine map from the factual state and requested edit to carrier coefficients. At test time, the predicted patch is applied once, followed by twelve autonomous transitions without future observations, teacher forcing, repeated hidden-state correction, or physical-state clamping.

For Single edits, rank 4 is the smallest rank on the preregistered grid that satisfies all development criteria. The frozen procedure replicates on independently trained checkpoints, remains effective at nearby anchors, and passes the preregistered specificity and integrity controls. We therefore interpret the rank-4 carrier as a compact, checkpoint-specific intervention interface for sustained counterfactual rollout—not as the model's intrinsic state dimension.

The same Single-derived carrier and Single-only affine map also support bounded same-object Joint requests, even though no Joint examples are used to fit the map. Across the matched training regimes, broader counterfactual support was associated mainly with better Joint rollout accuracy and more additive Joint hidden responses. This additive structure is partly, but not entirely, captured by the rank-4 subspace, and the effect of the patch quickly spreads through the rest of the hidden state. A position-edit stress test fails the required specificity controls. Together, these results support a compact intervention-entry interface rather than a closed four-dimensional dynamical state.

## 2 Related Work

Our work connects three lines of research: functional latent-state sufficiency, causal intervention in distributed representations, and physical structure in world and video models. Across these areas, the key distinction is between the information that can be decoded and the state that can be manipulated in a way that changes subsequent model dynamics. We focus on the latter: whether a compact hidden-state intervention can launch and sustain an autonomous counterfactual rollout.

### 2.1 Latent state, sufficiency, and action relevance

Latent-state learning seeks compact representations that preserve the information required for downstream behavior. DeepMDP formalized this idea by preserving reward and transition structure relevant to control (Gelada et al. 2019) , while action-sufficient representations extended the functional view to partially observable settings (Huang et al. 2022). Recent work argues more broadly that sufficiency depends on what the state must support—prediction, control, memory, planning, or intervention—rather than on whether its coordinates have an obvious interpretation (K.W. Kim 2026). Physically viable world-model formulations make the same point for intervention queries: similar passive observations may still require different internal states if they respond differently when perturbed (Thorpe et al. 2026).

Training objectives also affect how useful latent states become. Temporal prediction has been associated with more action-relevant representations than reconstruction alone (Yeom et al. 2026), and additional physical-state grounding can improve identifiability and planning in JEPA-style world models (Yan et al. 2026). Our main study does not introduce a new training objective. Instead, it asks whether an already trained recurrent world model contains a compact hidden-state intervention interface.

### 2.2 From probing to causal intervention in neural representations

Probing asks what information can be recovered from hidden representations. TCAV relates activation directions to human-defined concepts (B. Kim et al. 2018), but control-task analyses show that decodability alone does not establish that a model uses the recovered information (Hewitt and Liang 2019). Intervention methods therefore test function more directly. INLP, Amnesic Probing, and LEACE remove linearly represented information (Ravfogel et al. 2020; Elazar et al. 2021; Belrose et al. 2023), whereas Concept Bottleneck Models, activation engineering, and interchange interventions provide constructive ways to alter internal representations (Koh et al. 2020; Turner et al. 2023; Geiger et al. 2022). Distributed Alignment Search further allows causal variables to align with distributed subspaces rather than predefined coordinates (Geiger et al. 2024).

Causal edits must also be selective: they should change the targeted information without broadly corrupting unrelated state (Canby et al. 2025). We add a recurrent-world-model requirement. A one-shot hidden intervention must not merely alter an immediate readout; it must launch the intended autonomous trajectory and remain distinguishable from no-op, random, wrong-target, and structural reference conditions. The Joint extension additionally tests whether an interface learned from Single edits can support a bounded two-component request without fitting Joint examples in the addressable map.

### 2.3 Physical reasoning and object-centered dynamics

Physical prediction has motivated explicitly structured models. Interaction Networks represent objects and relations directly (Battaglia et al. 2016) , while later object-centric systems model long-horizon slot dynamics, separate object motion from interaction, or expose kinematic quantities more explicitly (Wu et al. 2022; Villar-Corrales, Wahdan, and Behnke 2023; Song et al. 2025). These approaches build useful structure into the model. Our question is different: after a recurrent world model has been trained without an explicit low-dimensional physical state, can a compact part of its distributed hidden state still serve as an intervention interface?

A complementary benchmark literature evaluates physical behavior rather than internal mechanism. IntPhys, CLEVRER, Physion, and IntPhys 2 test possible-versus-impossible events, collision reasoning, broader physical prediction, and object permanence or continuity (Riochet et al. 2018; Yi et al. 2019; Bear et al. 2021; Bordes et al. 2025). Representation-predictive video models can also acquire intuitive-physics behavior without an explicit physics engine (Garrido et al. 2025). Such results establish behavioral competence, but do not identify which hidden-state changes are used by the model's transition dynamics.

### 2.4 Physics representations, steering, and counterfactual consistency

The closest recent work directly examines physical representations in learned world and video models. Joseph et al. combine probing, subspace geometry, patch-level decoding, and ablations, and report that motion direction can be represented by a distributed population code rather than a single compact coordinate (Joseph et al. 2026). Zhang finds approximately linear game-state representations in two reinforcement-learning world models and shows that probe-derived hidden shifts can alter predictions (Zhang 2026), while Alam shows that a probe-derived Concept Activation Vector can steer physical-plausibility judgments in VideoMAE (Alam 2026). Together, these studies establish strong precedents for readable, distributed, and causally steerable physical representations.

Recent benchmarks also broaden physical evaluation beyond visual plausibility. CRONOS tests counterfactual consistency under controlled changes to scene factors (Begiristain, Dünkel, and Kortylewski 2026), while GAUGE evaluates physical fidelity against measurement-grounded observables and physical parameters (Wang et al. 2026). Our study targets a different level: hidden-state intervention. A compact, directly addressable edit is applied once, after which the recurrent model must autonomously continue the intended counterfactual future. We further test bounded compositional addressability and ask whether the low-rank carrier behaves as a self-contained dynamical state or as a compact entry interface whose effect subsequently propagates through the full hidden system.

## 3 Problem Formulation

This section defines the intervention problem and the notation used throughout the paper. We first consider bounded Single edits, which define the confirmatory rank-selection and replication assays, and then a same-object Joint extension. Single denotes an edit to one velocity component of one object; Joint denotes simultaneous edits to both velocity components of the same object. Formal definitions are collected in Appendix A, with environment, model, and training details in Appendix B.

### 3.1 World-model setting

We study two circular objects moving in a bounded, deterministic two-dimensional arena. Each object has a two-dimensional position and velocity, and the model receives noisy numeric object-state observations. All trajectories are generated with this simulator; no external physics dataset or pretrained trajectory dataset is used. This controlled setting gives an exact simulator-grounded counterfactual reference while leaving the learned hidden representation unconstrained.

For object j, the primitive object state consists of its horizontal and vertical positions and velocities. The joint physical state is the concatenation of the two object states:

$$\boldsymbol{s}_{j,t} = \left[x_{j,t},\, y_{j,t},\, v_{x,j,t},\, v_{y,j,t}\right]^{\mathsf{T}} \in \mathbb{R}^4, \qquad j \in \{A, B\}$$
$$\boldsymbol{s}_t = \left[\boldsymbol{s}_{A,t}^{\mathsf{T}},\, \boldsymbol{s}_{B,t}^{\mathsf{T}}\right]^{\mathsf{T}} \in \mathbb{R}^8 \tag{1}$$

In Eq. (1), $t$ is the discrete time index and $j \in \{A, B\}$ indexes the object. The vector $\boldsymbol{s}_{j,t} \in \mathrm{R}^4$ contains horizontal and vertical position $(x_{j,t}, y_{j,t})$ and velocity $(v_{x,j,t}, v_{y,j,t})$; $\boldsymbol{s}_t \in \mathrm{R}^8$ concatenates the two object states.

Derived quantities such as relative position, relative velocity, pair distance, and contact are used for evaluation but are not additional primitive coordinates. The world model maintains a 192-dimensional recurrent hidden state $\mathbf{z}_t$. After the anchor intervention, the model evolves autonomously from the patched state; no future observations are read.

The confirmatory carrier experiments use the factual + Single-counterfactual family, denoted MF1. Later matched comparisons also include MF0 (factual-only) and MF2 (factual + Single- and Joint-counterfactual support). Architecture and optimization budget are matched across families. Full specifications are in Appendix B.

### 3.2 Local physical counterfactual

Each factual/counterfactual pair shares the same pre-anchor history. In the primary Single family, the counterfactual changes exactly one velocity component of one object at the anchor; both positions and all other primitive velocity components remain fixed at that instant.

Let a unit edit direction specify the edited object and velocity axis. A local velocity change of $\delta v$ creates the counterfactual anchor state

$$\mathbf{s}_t^{\mathrm{CF}} = \mathbf{s}_t^{F} + \delta v\, \mathbf{e}_{j,a} \tag{2}$$

In Eq. (2), $\delta v$ is the signed requested velocity change, $j$ is the edited object, and $a \in \{x, y\}$ is the velocity axis. The selector $\boldsymbol{e}_{j,a} \in \mathbb{R}^8$ is zero except at the selected velocity coordinate. Superscripts F and CF denote the factual and native-counterfactual branches.

The later Joint extension uses the same physical counterfactual construction but changes both velocity components of the same object at the anchor. For object j, the Joint counterfactual $\boldsymbol{s}_t^{\mathrm{CF}}$ is

$$\boldsymbol{s}_t^{\mathrm{CF}} = \boldsymbol{s}_t^{\mathrm{F}} + \delta v_x\, \boldsymbol{e}_{\mathrm{j,x}} + \delta v_y\, \boldsymbol{e}_{\mathrm{j,y}}, \;\; \delta v_x \neq 0, \;\; \delta v_y \neq 0 \tag{2a}$$

For Joint requests, both velocity components of the same object are nonzero, while all positions and the other object's primitive state remain unchanged directly at the anchor. Joint edits remain within the registered magnitude and edited-anchor speed support and are evaluated separately from the original Single rank-selection procedure.

Later composition analyses use identity-matched $(F, S_x, S_y, J_{xy})$ quadruplets; exact matching and population definitions are given in Section 4.7 and Appendix H.

The edit is local only at the anchor. Algebraically derived quantities may change immediately, whereas later positions, contact timing, and post-collision velocities may change through simulator dynamics; only truly unaffected variables are required to remain stable. The native-counterfactual reference is obtained by passing the edited anchor observation through the model's normal update path, verifying that the model can propagate the requested counterfactual before low-rank reproduction is attempted. Details are in Appendix C.

### 3.3 Dynamics-effective latent intervention

We write a one-shot hidden-state intervention as:

$$\mathbf{z}_t^{*} = \mathbf{z}_t^{F} + \delta\mathbf{z} \tag{3}$$

In Eq. (3), $\boldsymbol{z}_t^{\mathrm{F}} \in \mathbb{R}^{192}$ is the recurrent state obtained from the factual history through anchor $t$; $\delta\boldsymbol{z} \in \mathbb{R}^{192}$ is the one-shot hidden perturbation; and $\boldsymbol{z}_t^{*}$ is the patched state. The superscript $*$ labels the intervention route and does not denote physical truth.

We call an intervention dynamics-effective when a single patch launches the intended counterfactual and the released model continues it autonomously while satisfying the registered target, preservation, physical-validity, and specificity criteria:

$$\mathrm{R}_H(\mathbf{z}_t^*) \approx \mathbf{Y}_{t:t+H}^{\mathrm{CF}} \tag{4}$$

In Eq. (4), $R_H$ is the autonomous rollout operator for $H$ future transitions, and $\boldsymbol{Y}_{t:t+H}^{\mathrm{CF}}$ is the simulator-grounded counterfactual reference trajectory. We use $H = 12$ in the main experiments. The relation $\approx$ denotes satisfaction of the registered multi-part evaluation, not coordinate-wise equality. Exact metrics and thresholds are given in Appendix F.

### 3.4 Low-rank and addressable latent carrier

The intervention becomes especially informative if the required hidden change can be restricted to a low-dimensional subspace. We represent a rank-r candidate carrier by a basis matrix and a coefficient vector:

$$\delta\mathbf{z} = \mathbf{U}_r\mathbf{c}, \qquad \mathbf{U}_r \in \mathbb{R}^{192\times r}, \quad \mathbf{c} \in \mathbb{R}^r \tag{5}$$

In Eq. (5), $r$ is the candidate carrier rank, $\boldsymbol{U}_r \in \mathbb{R}^{192\times r}$ has $r$ columns spanning the checkpoint-specific candidate carrier, and $\boldsymbol{c} \in \mathbb{R}^r$ is the intervention coefficient vector. Their product $\boldsymbol{U}_r\boldsymbol{c}$ is the 192-dimensional patch. The basis is obtained by the SVD construction in Section 4 and Appendix D; its individual columns are not assigned physical meanings.

A low-rank carrier is useful only if a new intervention can be specified without first computing the native counterfactual hidden state. We therefore require an addressable map that predicts the coefficient vector from the factual physical state and the requested edit:

$$\boldsymbol{c} = f_{\mathrm{addr}}(\boldsymbol{s}_t^{\mathrm{F}}, \boldsymbol{e}) \tag{6}$$

In Eq. (6), $f_{\mathrm{addr}}$ is the addressable coefficient map, $\boldsymbol{s}_t^{\mathrm{F}}$ is the factual two-object primitive state, and $\boldsymbol{e}$ is the requested velocity-edit vector. A Single request has one nonzero velocity slot; a Joint request has two nonzero slots for the same object. The output $\boldsymbol{c}$ determines the low-rank patch. At evaluation time, $f_{\mathrm{addr}}$ does not receive the native-counterfactual hidden state.

For a Joint request, the actual rollout uses the full affine prediction $\boldsymbol{c}_{\mathrm{full}} = f_{\mathrm{addr}}(\boldsymbol{s}_t^{\mathrm{F}}, \boldsymbol{e}_J)$. The zero-edit-subtracted component $\boldsymbol{c}_{\mathrm{request}} = f_{\mathrm{addr}}(\boldsymbol{s}_t^{\mathrm{F}}, \boldsymbol{e}_J) - f_{\mathrm{addr}}(\boldsymbol{s}_t^{\mathrm{F}}, \mathbf{0})$ is used only in the carrier-relative geometry analysis of Section 4.8 and Appendix I, not to generate the rollout intervention.

The corresponding test-time hidden intervention is then

$$\mathbf{z}_t^* = \mathbf{z}_t^F + \mathbf{U}_r\mathbf{c} \tag{7}$$

Eq. (7) applies $\boldsymbol{U}_r\boldsymbol{c}$ once to $\boldsymbol{z}_t^{\mathrm{F}}$ to form $\boldsymbol{z}_t^*$, after which the addressable map is not consulted again. In the Joint extension, the rank-4 carrier and affine map remain Single-derived and Single-only fitted: no Joint hidden differences or Joint map examples are used to construct the primary addressable Joint patch.

### 3.5 Operational domain and claim boundary

The intervention claims in this paper are defined over finite operational domains rather than as global properties of the model. We use the same generic decomposition throughout:

$$\Omega = \Omega_{state} \times \Omega_{int} \times \Omega_{time} \tag{8}$$

In Eq. (8), $\Omega_{\text{state}}$ is the tested factual-state region, $\Omega_{\text{int}}$ the registered intervention family and magnitude support, and $\Omega_{\text{time}}$ the evaluated anchor times; $\times$ denotes a Cartesian product. The primary Single domain uses bounded one-object, one-component edits, with rank selection and replication at $t = 7$ and temporal tests at $t = 5\text{–}9$. The Joint extension is evaluated at $t = 7$ under the matched support defined in Appendix C.

These bounds also limit the interpretation of the experiment results. A passing rank-4 carrier does not imply four named physical variables, an intrinsic state dimension of four, or a closed four-dimensional dynamical state. Successful Joint editing establishes only bounded same-object compositional addressability. The matched training-regime comparison is descriptive and does not imply a universal relation between training objective and latent structure.

# 4 Method

This section describes the primary Single-intervention workflow and the later Joint and mechanistic extensions. The primary workflow first verifies native counterfactual adequacy, fits a checkpoint-specific low-rank carrier from training-only hidden differences, learns an addressable coefficient map, applies one hidden patch, and evaluates the released autonomous rollout with registered controls. Rank selection and fresh-checkpoint evaluation use separate data roles. The selected rank and Single-derived construction are then kept fixed for the Joint analyses; the later extensions do not reopen the original rank decision.

## 4.1 Native counterfactual adequacy evaluations

Before interpreting any low-rank result, we verify that the model can produce the requested counterfactual when the edited anchor state is supplied through its normal observation/update path. For intervention unit $i$ at anchor $t$,

$$\Delta\mathbf{z}_i^{\text{full}} = \mathbf{z}_{t,i}^{\text{CF}} - \mathbf{z}_{t,i}^{\text{F}}$$
$$\mathbf{z}_{t,i}^{\text{F}} + \Delta\mathbf{z}_i^{\text{full}} = \mathbf{z}_{t,i}^{\text{CF}} \tag{9}$$

We evaluate three reference routes. G0 releases the factual anchor state. G1 takes the edited anchor observation, forming $\mathbf{z}_t^*$, and then releases the resulting native counterfactual state. G2 adds the complete native counterfactual-minus-factual hidden difference to the factual state, as shown in Eq. (9). G1 is the model-adequacy reference and G2 checks the hidden-injection interface. Both must pass before a low-rank failure or success is interpreted. The same logic is used for the later Joint extension. Detailed route and gate definitions are given in Appendices C, F, and G.

## 4.2 Constructing candidate low-rank carriers

Carrier construction is checkpoint-specific and uses only the carrier-fit split. For each primary Single unit, we compute the native counterfactual-minus-factual anchor difference. At $t = 7$, the fit matrix contains $N_{\text{fit}} = 1024$ differences in the 192-dimensional hidden space:

$$\mathbf{D} = \begin{bmatrix} \Delta\mathbf{z}_1^\top \\ \Delta\mathbf{z}_2^\top \\ \vdots \\ \Delta\mathbf{z}_{N_{\text{fit}}}^\top \end{bmatrix} \in \mathbb{R}^{N_{\text{fit}}\times 192} \qquad N_{\text{fit}} = 1024 \tag{10}$$

We use the uncentered SVD of the raw difference matrix,

$$\mathbf{D} = \mathbf{L\Sigma V}^\top$$
$$\mathbf{U}_r = \mathbf{V}_{:,1:r} \in \mathbb{R}^{192\times r}$$

(11)

The columns of $\mathbf{U}_r$ span the rank-$r$ candidate carrier; they are not assigned physical meanings. For a privileged capacity analysis, the native hidden difference of unit $i$ can be projected onto the carrier,

$$\mathbf{c}_i^{\text{oracle}} = \mathbf{U}_r^\top \Delta \mathbf{z}_i$$

(12)

These oracle coefficients $\mathbf{c}^{\text{oracle}}$ test subspace capacity but are not used by the primary addressable intervention. The Joint extension keeps the selected rank fixed at four and uses checkpoint-specific carriers built from Single fit differences only; Joint hidden differences are used only for privileged capacity references. MF0 is therefore included in native Joint and composition analyses, whereas Single-derived carrier/addressability comparisons are defined for MF1 and MF2. Appendix D gives the full carrier construction.

### 4.3 Addressable coefficient map

The test-time intervention must be specified without the evaluation unit's native counterfactual hidden state. We therefore fit, on Single carrier-fit units only, an affine ridge map from the factual primitive state and requested velocity edit to carrier coefficients. Let $\mathbf{e}_i \in \mathbb{R}^4$ encode the requested velocity edit and $\mathbf{s}_{t,i}^{\text{F}} \in \mathbb{R}^8$ the factual two-object primitive state. Define

$$\mathbf{x}_i = \begin{bmatrix} \mathbf{e}_i \\ \mathbf{s}_{t,i}^{\text{F}} \end{bmatrix} \in \mathbb{R}^{12}$$
$$\tilde{x}_{ik} = \frac{x_{ik} - \mu_k}{\max(\sigma_k, 10^{-6})}, \qquad k = 1, \ldots, 12$$
$$\mathbf{c}_i = f_{\text{addr}}\left(\mathbf{s}_{t,i}^{\text{F}}, \mathbf{e}_i\right) = \boldsymbol{W}_r^\top \tilde{\mathbf{x}}_i + \mathbf{b}_r$$

(13)

The feature statistics $\mu_k$ and $\sigma_k$, coefficient matrix $\mathbf{W}_r$, and intercept $\mathbf{b}_r$ are fitted on the carrier-fit split only. The fitting objective and regularization setting are given in Appendix E. At evaluation time, $f_{\text{addr}}$ receives only the factual state, requested edit, and frozen fit statistics.

For a Joint request, the same feature representation is used with two nonzero edit components for the same object. The rank-4 map remains Single-only fitted for both MF1 and MF2; no Joint example is used to refit its standardization or regression parameters. The actual Joint rollout uses the full affine prediction $\mathbf{c}_{\text{full}} = f_{\text{addr}}\left(\mathbf{s}_t^{\text{F}}, \mathbf{e}_{xy}\right)$. The zero-edit-subtracted component $\mathbf{c}_{\text{request}} = f_{\text{addr}}\left(\mathbf{s}_t^{\text{F}}, \mathbf{e}_{xy}\right) - f_{\text{addr}}\left(\mathbf{s}_t^{\text{F}}, \mathbf{0}\right)$ is used only in the later geometry analysis. Appendix E gives the complete fitting objective and affine-composition identity.

### 4.4 One-shot autonomous rollout

The predicted coefficient vector is mapped back to the full hidden space and added once at the anchor:

$$\mathbf{z}_{t,i}^* = \mathbf{z}_{t,i}^{\text{F}} + \beta \mathbf{U}_r \mathbf{c}_i.$$

(14)

The primary Single and later Joint addressable routes use $\beta = 1$. After the patch, the model runs for 12 autonomous transitions, producing 13 decoded states including the anchor. There are no future observation reads, teacher forcing, repeated hidden-state edits, physical/hidden clamping, future oracle states, or decoder feedback; all evaluated actions are zero.

The primary Single route is scored by the registered scientific metrics and autonomy/eligibility gates in Appendix F. The Joint extension uses the same one-shot rollout, with primarily the direct-target and unaffected-slot definitions adapted to the two-component request. Native Joint, $\mathbf{U}_4$-oracle, and Single-only addressable routes are compared on the same held-out population defined in Appendix H.

### 4.5 Controls, rank selection, and fresh replication

For Single edits, controls include no-patch/sham, independent random equal-norm perturbations, wrong-object and wrong-time edits, full-hidden references, and leakage/identity checks. The Joint extension uses matched no-patch, random equal-norm, wrong-object, and wrong-vector controls. Exact route definitions and aggregation are given in Appendices G and H.

Rank is selected only for the primary Single experiment on three development checkpoints. The registered grid is $\mathcal{R} = \{1,2,4,8,12,16,20,32\}$, with rank 64 diagnostic, rank 0 a no-patch reference, and rank 192 the full-hidden reference. We evaluate the Single intervention in two preregistered trajectory strata: S1, where neither the factual nor counterfactual branch contains a detected contact step over the 12-transition horizon, and S2, where the factual branch contains a detected contact step and the edit produces a sufficiently large post-anchor factual–counterfactual trajectory difference. The exact programmatic definitions are given in Appendix C.4 and Appendix F.7. S1 and S2 are evaluated separately and are never pooled. The selected rank is

$$r^\star = \min\left\{ r \in \mathcal{R}\colon \sum_{q=1}^{3} \mathbb{1}\left[\mathrm{Pass}_{q,r,\mathrm{S1}} \wedge \mathrm{Pass}_{q,r,\mathrm{S2}}\right] \geq 2 \right\}. \tag{15}$$

Here $q$ indexes the three independently trained development checkpoints, and $\mathbb{1}[\cdot]$ is the indicator function. Thus $r^\star$ is the smallest tested rank passing both S1 and S2 strata in at least two of three checkpoints.

After selection, the rank, $\beta$, feature definition, standardization, ridge setting, thresholds, and panel rule are frozen. Three fresh checkpoints were trained independently from scratch using the same architecture, dataset, and frozen training procedure as the development checkpoints, but with new random seeds; none was used during method development or rank selection. Each fresh checkpoint fits its own carrier and addressable map using the frozen procedure and its own carrier-fit split; the confirmatory split is evaluation-only. No new rank sweep or retuning is allowed. The later Joint extension reuses this selected rank and Single-derived construction but does not participate in the original rank-selection or fresh-replication decision.

### 4.6 Temporal tests

The temporal assays characterize the established Single rank-4 interface. B1 refits the same frozen procedure independently at each $t \in \{5,6,7,8,9\}$ and evaluates it at the same anchor. B2 transports the exact $t = 7$ basis, affine-map parameters, feature statistics, rank, and $\beta$ to $t = 5,6,8,9$ without refitting or restandardization. These assays test temporal recovery and reuse of the Single interface only; the Joint extension is evaluated at $t = 7$.

### 4.7 Matched Joint-response and composition analyses

Each Joint composition unit is an identity-matched quadruplet: factual $F_i$, horizontal Single $S_{x,i}$, vertical Single $S_{y,i}$, and Joint $J_{xy,i}$. The four routes share the same base unit, edited object, anchor $t = 7$, pre-anchor history, and realized observation noise, and the two Single edit vectors sum to the Joint request. The registry contains 512 unique units, split into 411 FIT units for factual-reference whitening/local-operator fitting and 101 held-out TEST units for composition and route comparisons.

With the frozen checkpoint-specific whitening map $\mathrm{wh}(\cdot)$, define

$$
\begin{aligned}
\Delta \mathbf{z}_{x,i}^{\mathrm{wh}} &= \mathrm{wh}\left(\mathbf{z}_{S_x,i}\right) - \mathrm{wh}\left(\mathbf{z}_{F,i}\right) \\
\Delta \mathbf{z}_{y,i}^{\mathrm{wh}} &= \mathrm{wh}\left(\mathbf{z}_{S_y,i}\right) - \mathrm{wh}\left(\mathbf{z}_{F,i}\right) \\
\Delta \mathbf{z}_{xy,i}^{\mathrm{wh}} &= \mathrm{wh}\left(\mathbf{z}_{J_{xy},i}\right) - \mathrm{wh}\left(\mathbf{z}_{F,i}\right) \\
\Delta \mathbf{z}_{\mathrm{add},i}^{\mathrm{wh}} &= \Delta \mathbf{z}_{x,i}^{\mathrm{wh}} + \Delta \mathbf{z}_{y,i}^{\mathrm{wh}}
\end{aligned}
\tag{16}
$$

The primary static composition error is

$$
E_{\mathrm{add},i} = \frac{2\left\|\Delta \mathbf{z}_{xy,i}^{\mathrm{wh}} - \Delta \mathbf{z}_{\mathrm{add},i}^{\mathrm{wh}}\right\|_2}{\left\|\Delta \mathbf{z}_{xy,i}^{\mathrm{wh}}\right\|_2 + \left\|\Delta \mathbf{z}_{\mathrm{add},i}^{\mathrm{wh}}\right\|_2 + \varepsilon} \qquad \varepsilon = 10^{-12}
\tag{17}
$$

Lower $E_{\mathrm{add},i}$ indicates a more additive native Joint hidden response. Dynamic composition additionally propagates the native Joint and additive initializations through the actual recurrent model at $k \in \{0,1,3,6,12\}$; the full definition is in Appendix H.

For MF1 and MF2, Native Joint, privileged $\mathbf{U}_4$-oracle, and Single-only addressable routes are compared on the same 101 TEST units per checkpoint, together with matched specificity controls. MF0 participates in native Joint and composition analyses but not in $\mathbf{U}_4$-oracle/addressable comparisons. Checkpoints, not units, are the replication level; family summaries are descriptive medians over three checkpoints.

## 4.8 Carrier-relative and recurrent-dynamics analyses

For mechanistic analysis, each checkpoint-specific Single-derived $\mathbf{U}_4$ is expressed in the factual-reference whitened coordinates from Section 4.7. Let $\mathbf{W}_{\mathrm{wh}}$ be the whitening matrix and

$$
\begin{aligned}
\mathbf{Q}_U &= \mathrm{orth}(\mathbf{W}_{\mathrm{wh}}\mathbf{U}_4) \\
\mathbf{P} &= \mathbf{Q}_U\mathbf{Q}_U^{\top} \\
\mathbf{Q} &= \mathbf{I}_{192} - \mathbf{P}
\end{aligned}
\tag{18}
$$

$\mathbf{P}$ projects onto the whitened Single-derived carrier and $\mathbf{Q}$ onto its orthogonal complement. We use these projectors to separate composition error and squared-norm concentration inside and outside the carrier; these are descriptive diagnostics and do not assume that composition is confined to $\mathbf{U}_4$.

We also fit checkpoint- and route-specific local affine recurrent operators $\mathbf{K}_F$, $\mathbf{K}_S$, and $\mathbf{K}_J$ on consecutive whitened hidden states from the 411 FIT units and evaluate them on TEST units. The Single-to-Joint operator shift is

$$
\begin{aligned}
\Delta \mathbf{K}_{JS} &= \mathbf{K}_J - \mathbf{K}_S, \\
d_{JS} &= \frac{\left\|\Delta \mathbf{K}_{JS}\right\|_F}{\left\|\mathbf{K}_F\right\|_F + \varepsilon}.
\end{aligned}
\tag{19}
$$

The operators and $\Delta\mathbf{K}_{JS}$ are further decomposed into $\mathbf{P}/\mathbf{Q}$ blocks to quantify carrier retention, leakage to the complement, reverse transfer, and dimension-normalized block strength. These are local descriptive fits, not a global Koopman representation or a closed reduced-order model.

Finally, the Single-only affine Joint prediction is decomposed into full, zero-edit, and request-only components. The actual rollout always uses the full affine patch; distances to the projected native Joint oracle are geometry diagnostics only. Appendix I gives the complete projector, operator, block, enrichment, and patch-geometry definitions.

### 4.9 Position-edit stress test

A separate development-only stress test asks whether the evaluation distinguishes target-specific intervention from generic hidden sensitivity. At $t = 7$, one $x$ or $y$ position coordinate of one object is shifted by 0.05, 0.10, or 0.20; all velocities and the remaining primitive anchor coordinates are held fixed, and simulator counterfactual truth is regenerated from the edited state.

The intended low-rank position patch is compared with no-patch, independent random equal-norm, and wrong-object alternatives. Raw rollout improvement is not sufficient if unrelated or incorrectly targeted controls produce comparable behavior. This diagnostic is separate from the primary velocity rank selection and confirmatory replication; details are given in Appendices C and G.

## 5 Results

The results follow the evidential order: native counterfactual adequacy, development-stage rank selection, fresh-checkpoint replication, temporal reuse, bounded Joint addressability, matched training-family comparisons, mechanistic diagnostics, and the position-edit negative contrast.

### 5.1 The model natively supports the requested counterfactual rollouts

Before interpreting any low-rank intervention, we verified that the trained model could autonomously propagate the requested Single counterfactual when shown the complete edited anchor condition. Across the three development checkpoints, the factual (G0), native-counterfactual (G1), and full-hidden-equivalence (G2) routes passed all required S1/S2 tests.

The same prerequisite held on three independently trained fresh checkpoints. All 18 route-by-stratum tests passed, and the lowest unit-level joint-pass coverage in any G0/G1 test was 0.922, above the registered 0.80 requirement. G2 also reconstructed the native counterfactual hidden state and rollout within numerical tolerance for all 1536 fresh-checkpoint unit evaluations.

Thus the model and injection interface can represent and propagate the requested counterfactual before any low-rank claim is made.

### 5.2 A rank-4 addressable carrier is the smallest passing tested rank

We next evaluated the preregistered rank grid on the development panel using the complete addressable procedure. Ranks 1 and 2 failed the registered panel rule, whereas rank 4 was the smallest tested rank that passed both S1 and S2 on at least two of the three development checkpoints together with the mandatory controls (as shown in Figure 1a). Higher tested ranks could also pass, but the rule selects the smallest passing tested rank.

A privileged oracle-projection sweep identified the same smallest passing tested rank, showing that the carrier itself had sufficient capacity when the correct coordinates were supplied. The addressable rollout remains the primary evidence because the oracle route uses the evaluation unit's native counterfactual hidden state. The hidden-difference spectrum in Figure 1b is therefore descriptive: it suggests compressibility but was not used as the rank-selection criterion.

Accordingly, rank 4 is the smallest tested rank sufficient for the registered Single intervention family under this assay. It is not an estimate of the model's intrinsic state dimension, a unique basis, or a proof that the full hidden response is exactly four-dimensional.

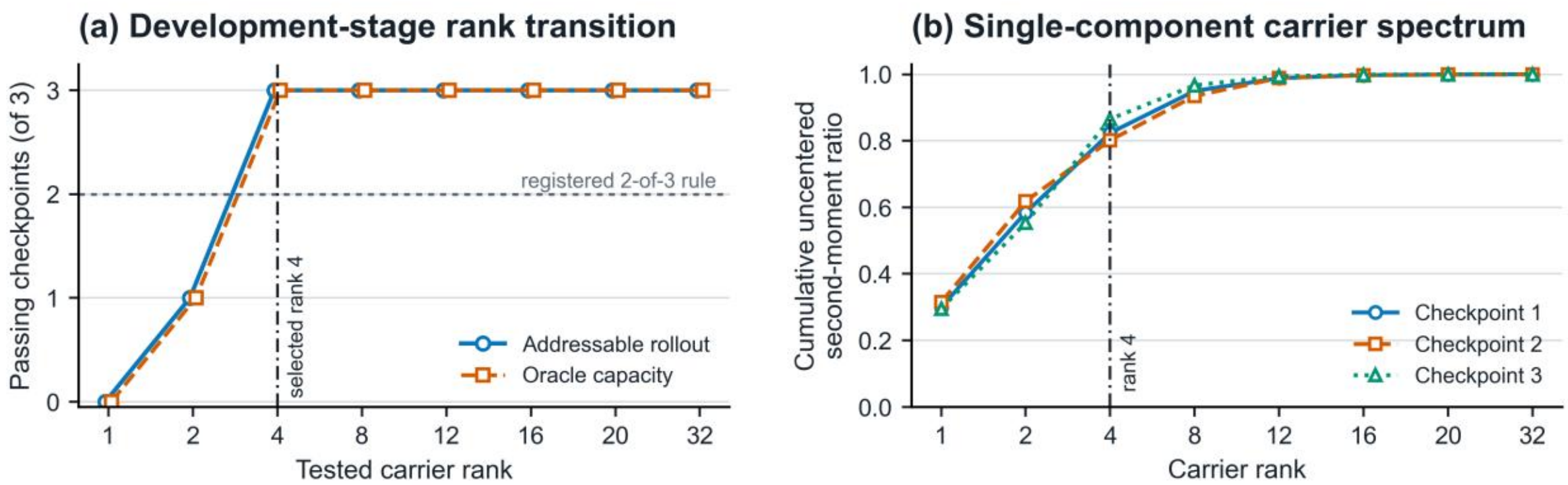

Figure 1. Development-stage rank transition and carrier geometry for the primary Single experiment. (a) Addressable-rollout rank sweep; rank 4 is the smallest tested rank satisfying the registered panel rule and controls. The privileged oracle-projection sweep identifies the same smallest passing tested rank. (b) Descriptive hidden-difference spectrum and cumulative uncentered second-moment ratios. Rank selection is based on rollout behavior and controls, not on a spectral threshold.

### 5.3 The frozen rank-4 procedure replicates under the fresh-checkpoint rule

After rank selection, the rank, intervention strength, feature definition, standardization, ridge setting, thresholds, and panel rule were frozen before evaluation on separately trained fresh checkpoints. Each fresh checkpoint fitted its own carrier and addressable map only on its carrier-fit split; no rank sweep or post-hoc retuning was allowed.

The frozen procedure satisfied the preregistered 2-of-3 fresh-checkpoint rule (as shown in Figure 2a). Unit-level joint-pass coverage in S1/S2 was 0.867/0.914 for Fresh 1, 1.000/0.922 for Fresh 2, and 0.711/0.973 for Fresh 3. Fresh 1 and Fresh 2 therefore passed both strata; Fresh 3 missed only S1. All six fresh model-by-stratum cells passed the registered median, numerical-eligibility, and autonomy criteria, so the single miss was specifically a joint-pass-coverage failure.

The registered specificity controls also behaved as required (as shown in Figure 2b). Sham/rank-zero, wrong-object, wrong-time, leakage, and identity/order checks passed, while the independent random equal-norm perturbation produced 0/3 fresh-model target passes. The result therefore supports bounded replication of the frozen procedure, with checkpoint-level heterogeneity rather than uniform success across every test.

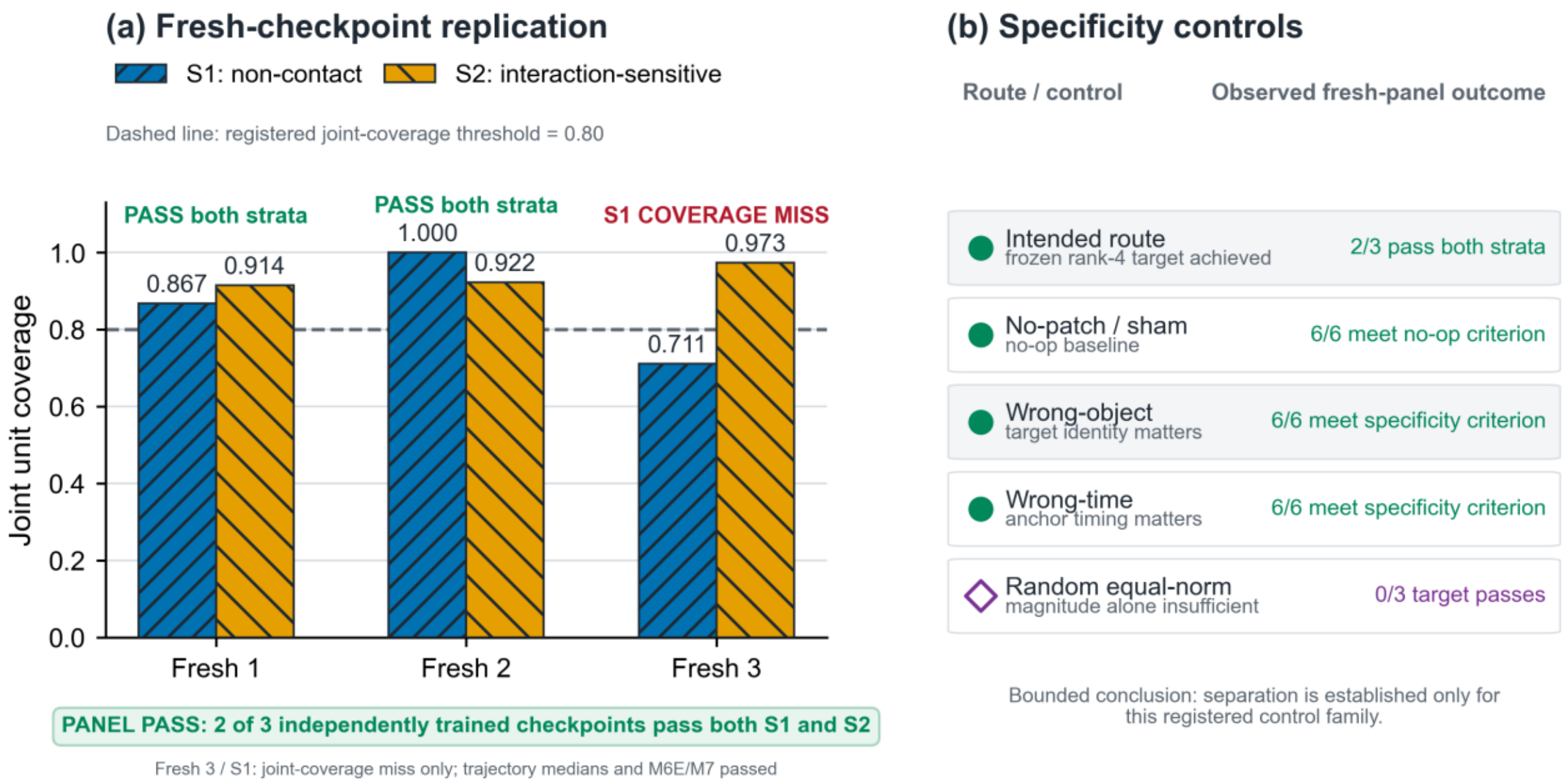

Figure 2. Fresh-checkpoint replication and specificity controls for the frozen rank-4 addressable carrier. (a) Unit-level joint-pass coverage in S1 and S2; Fresh 1 and Fresh 2 pass both strata, whereas Fresh 3 misses only S1, satisfying the registered 2-of-3 rule. (b) Registered reference and specificity-control outcomes; a passing control means that it behaves according to its own registered criterion.

### 5.4 The rank-4 interface can be recovered and reused across nearby anchors

We separated temporal generalization into recoverability and direct reuse. B1 refitted a checkpoint-specific rank-4 interface independently at each anchor from $t = 5$ to $t = 9$ while keeping the selected rank, map family, intervention strength, thresholds, and panel rule fixed. At every tested anchor, two of the three fresh checkpoints passed both S1 and S2 (as shown in Figure 3a).

B2 tested the stronger condition: the exact interface fitted at $t = 7$, i.e. the carrier basis, ridge parameters, feature statistics, rank, and intervention strength, was transported unchanged to $t = 5,6,8,9$. All four target anchors again satisfied the registered 2-of-3 rule, with no target-anchor refitting or restandardization (as shown in Figure 3b).

These tests support local temporal reuse over the tested $t = 5$–$9$ window. They do not establish globally time-invariant coordinates, a closed low-dimensional Markov state, or temporal transport of the later Joint extension.

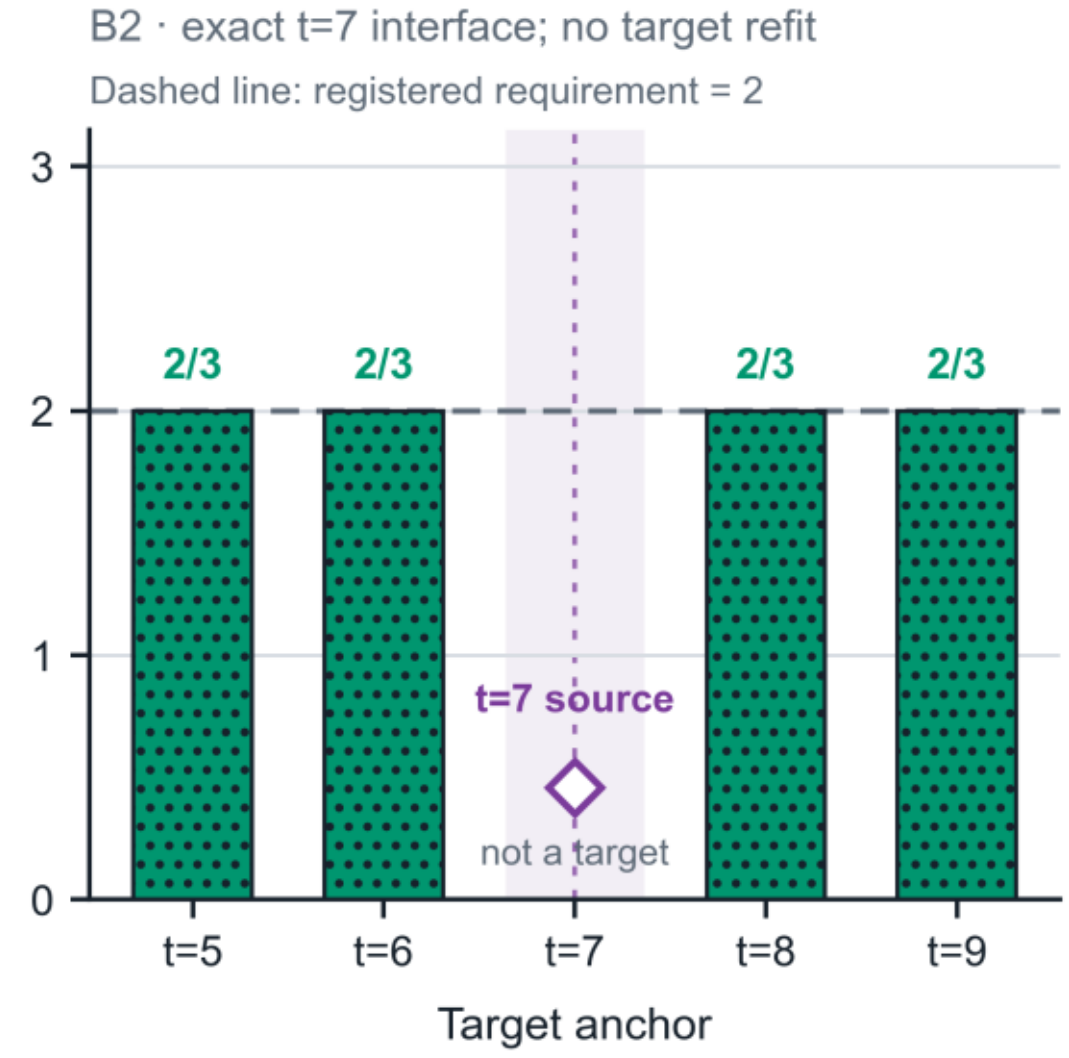

Figure 3. Temporal recoverability and direct reuse of the primary Single rank-4 interface. (a) B1 refits a checkpoint-specific interface at each anchor from t=5 to t=9. (b) B2 transports the exact t=7 interface to t=5,6,8,9 without refitting or restandardization; the t=7 marker denotes the source interface.

### 5.5 A Single-derived rank-4 interface supports bounded same-object Joint requests

We next asked whether the Single-derived interface could support a same-object Joint request in which both velocity components are edited simultaneously. Three routes were compared on the same 101 held-out units per checkpoint: Native Joint, a privileged projection of the native Joint hidden difference onto the frozen Single-derived $\mathbf{U}_4$, and the stricter addressable route generated by the Single-only affine map without using the native Joint hidden state at test time.

The addressable route remained successful in both intervention-trained families, as shown in Figure 4a. Family-median joint-pass coverage for Native/$\mathbf{U}_4$-oracle/Addressable was 0.970/0.941/0.901 in MF1 and 0.990/1.000/0.980 in MF2. The corresponding addressable rollout RMSE was 0.157 in MF1 and 0.125 in MF2. Matched sham/no-patch, random equal-norm, wrong-object, and wrong-vector controls did not reproduce the intended effect as shown in Figure 4b. Thus the Single-derived carrier and Single-only map support bounded same-object Joint addressability. Still, this does not establish a universal additive algebra, arbitrary multi-component control, or an object-independent intervention law.

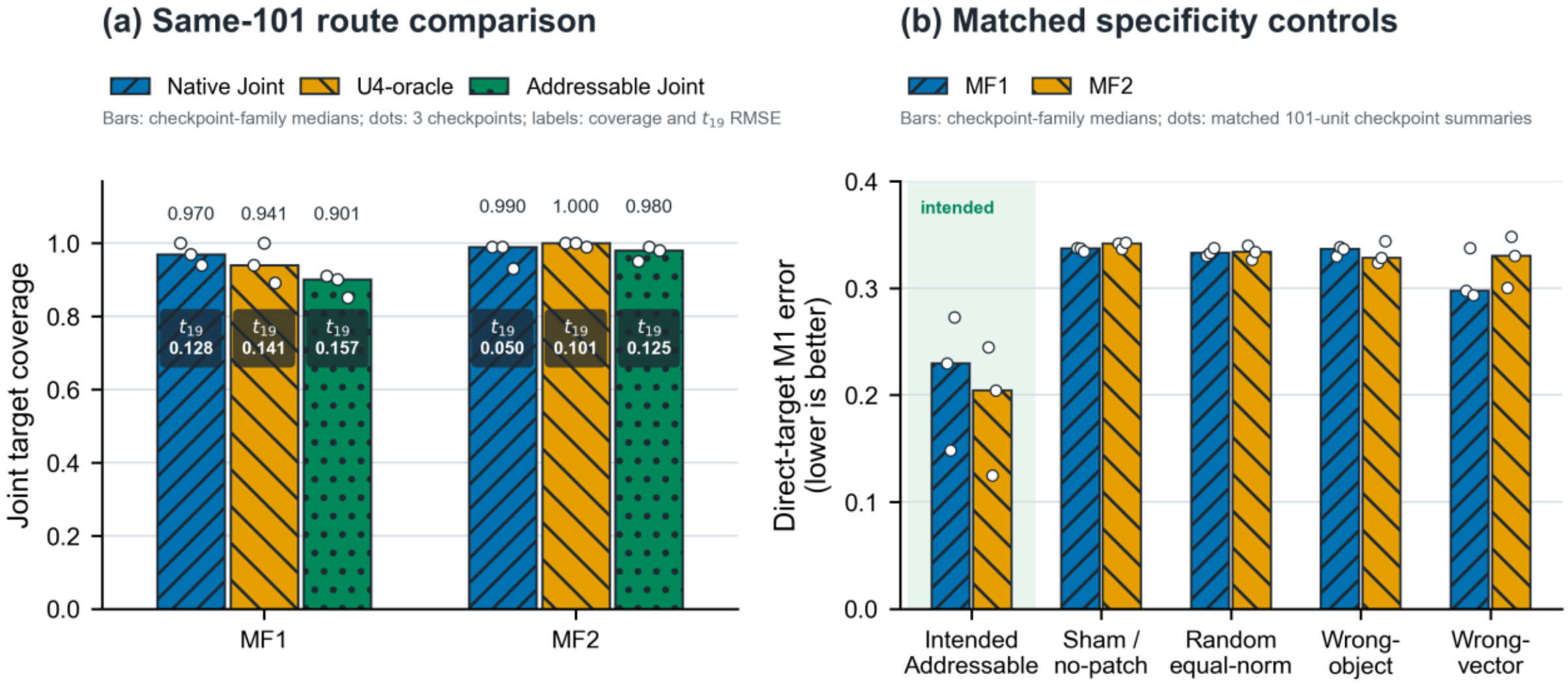

Figure 4. Bounded same-object Joint addressability from a Single-derived rank-4 interface. (a) Native Joint, $\mathbf{U_4}$-oracle, and Single-only addressable routes on the same held-out units in MF1 and MF2. Bars show family medians and dots show checkpoint-level values. (b) Matched specificity controls for the addressable Joint route; lower direct-target error is better.

## 5.6 Broader intervention support is associated with more accurate and more compositional Joint responses

We then compared native Joint behavior across the matched MF0, MF1, and MF2 training families. Native Joint competence is already substantial in MF0 and even stronger in MF1 and MF2: family-median joint-pass coverage is 0.881, 0.957, and 0.961 for MF0, MF1, and MF2, respectively (as shown in Figure 5a). The larger change is rollout fidelity, with median terminal RMSE decreasing from 0.179 to 0.130 to 0.050. Broader intervention support is therefore associated mainly with more accurate Joint rollout rather than a transition from failure to success.

The native hidden responses also shift toward additivity, as shown in Figure 5b. Family-median static composition error $E_{\text{add}}$ decreases from 0.858 in MF0 to 0.624 in MF1 and 0.495 in MF2. The shift is clear across model families, but not at every individual checkpoint. Dynamic composition gives weaker supporting evidence: at $k = 12$, the family-median hidden-space error is approximately 0.293, 0.229, and 0.218. We therefore treat the static result as the primary composition evidence and the dynamic result as a secondary diagnostic.

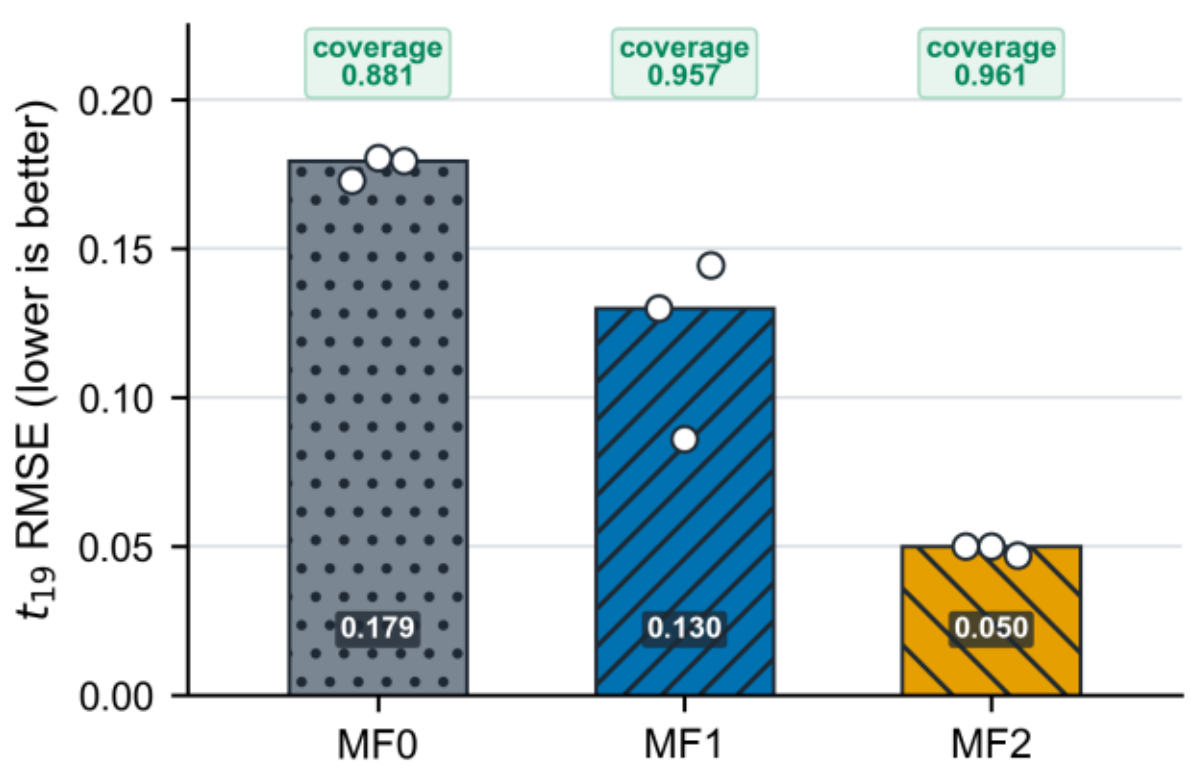

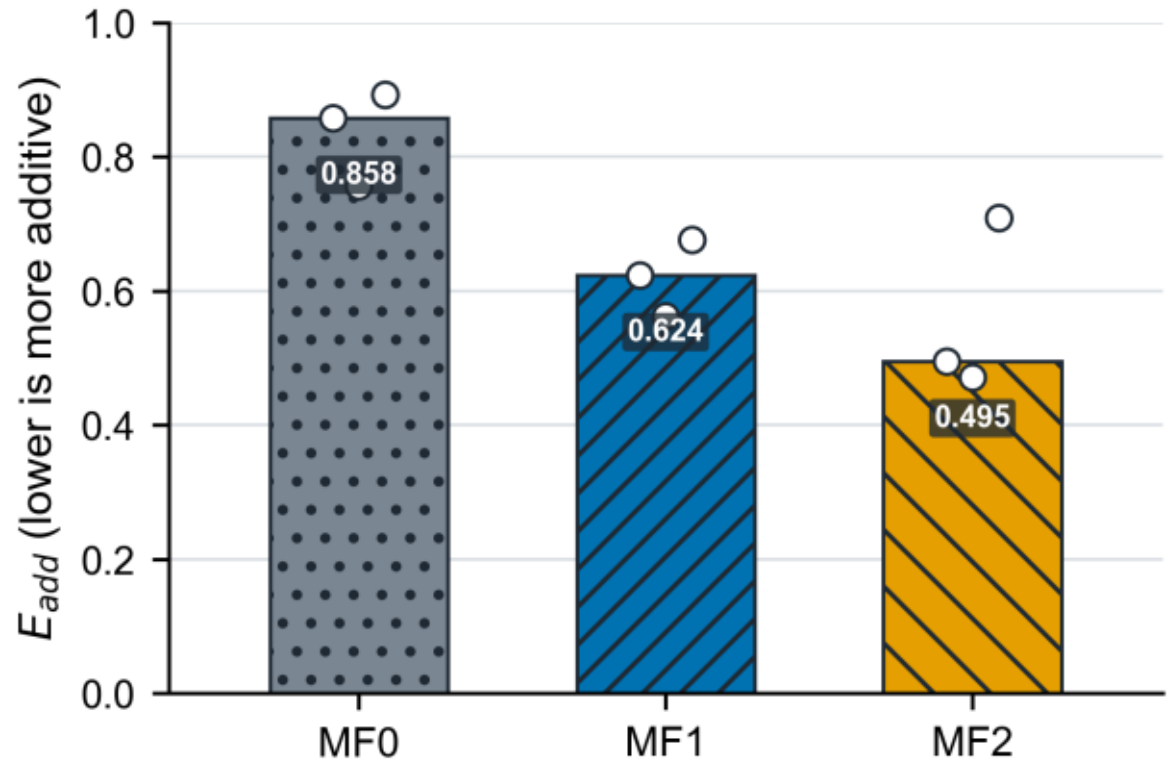

Figure 5. Native Joint rollout fidelity and static hidden-response composition across matched training families. (a) Native Joint terminal RMSE, with family-median joint-pass coverage reported above the bars. (b) Static composition error $\boldsymbol{E}_{\mathbf{add}}$; lower values indicate closer agreement between the native Joint response and the additive combination of matched Single responses. Dots show checkpoint-level values.

### 5.7 The rank-4 carrier is a compact entry interface into high-dimensional recurrent dynamics

Carrier-relative analyses clarify what the rank-4 result does and does not represent. Composition is not confined to the Single-derived $\mathbf{U}_4$. In Figure 6a, the family-median static composition error of MF1 is 0.655 inside the carrier and 0.606 in its orthogonal complement; in MF2, the corresponding values are 0.498 and 0.514. The shift toward greater additivity therefore occurs in both regions. Nevertheless, $\mathbf{U}_4$ is strongly enriched: although it occupies only 4 of 192 whitened dimensions, it contains 30.4% of the composition-defect squared norm in MF1 and 17.5% in MF2, about 14.6-fold and 8.4-fold above the dimensional baseline.

The fitted local recurrent operators show that the rank-4 entry is not approximately closed, as shown in Figure 6b. Family-median one-step carrier retention $R_{\text{in}}$ is 0.016 for both Single and Joint in MF1 and 0.027/0.028 in MF2; the complementary leakage is therefore 0.972–0.984. Single and Joint recurrence also remain distinguishable: the normalized operator difference $d_{JS}$ is 1.330 in MF1 and 1.667 in MF2.

A separate geometry diagnostic gives another boundary. The full addressable Joint patch is not closer to the projected native-Joint oracle in MF2 than in MF1: family-median symmetric distance is 0.770 versus 0.631, despite better MF2 Joint behavior. This does not make geometry irrelevant; it only shows that distance to this particular oracle projection is not a sufficient proxy for dynamics-effective addressability.

Together, these analyses support a compact intervention-entry interpretation. The rank-4 carrier concentrates intervention- and composition-relevant structure and is sufficient to launch the requested future, but the perturbation is then propagated through the full recurrent hidden state rather than remaining in a closed four-dimensional subsystem.

Figure 6. Carrier-relative composition and recurrent unfolding of the Single-derived rank-4 interface. (a) Static composition error inside the whitened $\mathbf{U}_4$ carrier and its orthogonal complement; annotations report composition-defect fraction and enrichment relative to the 4/192-dimensional baseline. (b) One-step carrier retention $R_{\text{in}}$ for the fitted local Single and Joint operators; low retention implies high complementary leakage.

## 5.8 Position edits do not satisfy the dynamics-effective evidence standard

Finally, we use a development-only position-edit stress test as a negative contrast. At $t = 7$, one position coordinate is shifted by 0.05, 0.10, or 0.20 while all velocity components and the remaining primitive anchor coordinates are fixed; simulator counterfactual truth is regenerated from the edited state.

The raw rollout gate is not specific to the intended patch as shown in Figure 7a. At every amplitude, the intended position patch, rank-zero/no-patch route, and independent random equal-norm perturbation all pass in 3/3 development checkpoints. Figure 7b reveals that the specificity test is more discriminating: wrong-object specificity and the full specificity criterion fail in 0/3 checkpoints at every amplitude.

The tested position intervention therefore does not meet the dynamics-effective evidence standard. This does not imply that position information is absent or that no position intervention could succeed; it shows that raw outcome improvement alone is insufficient without separation from no-patch, random, and wrong-target alternatives.

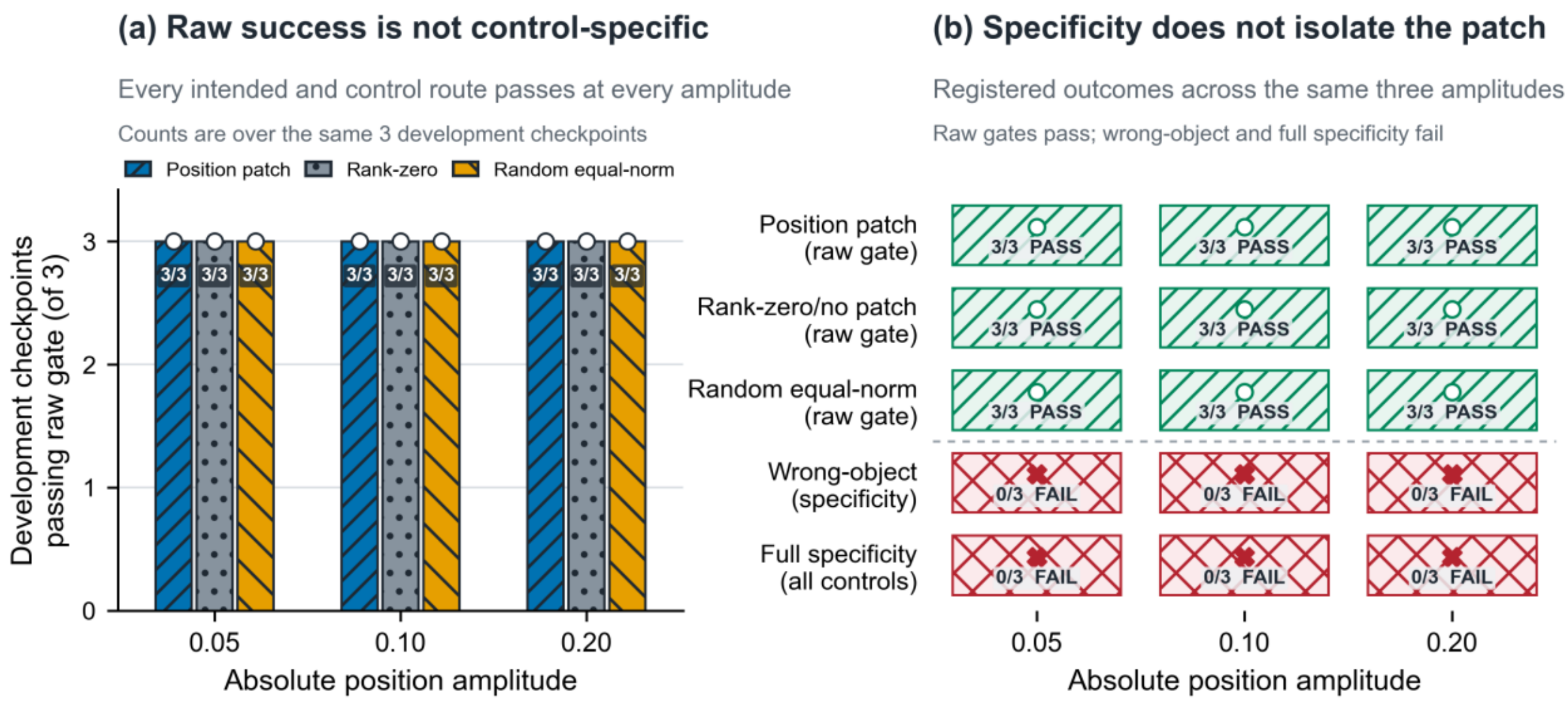

Figure 7. Position-edit stress test. (a) The intended position patch, no-patch route, and random equal-norm route all satisfy the raw rollout gate in 3/3 development checkpoints at every amplitude. (b) Wrong-object specificity and the full specificity criterion fail in 0/3 checkpoints at every amplitude, so the raw gate does not isolate the intended intervention.

# 6 Discussion

The results support a compact, dynamics-effective intervention interface for the tested velocity counterfactuals. The primary Single experiments identify a checkpoint-specific rank-4 interface that can be addressed from the factual state, patched once, and propagated through an autonomous future. The later Joint extension shows bounded compositional addressability with the same Single-derived carrier and Single-only map. Mechanistic analyses qualify this result: the compact entry interface is not a closed four-dimensional dynamical state, because its effect rapidly couples to the larger recurrent hidden system.

## 6.1 Dynamics-effective intervention is a functional, non-semantic notion

Our criterion is functional. Decodability or an immediate steering effect does not show that an edited representation can support the intended future. We therefore intervene once and evaluate what the model does afterward: the patch must be addressable from the factual state and requested edit, survive twelve autonomous transitions, and remain separated from the registered negative and integrity controls.

This view is consistent with functional notions of latent-state sufficiency, where representations are judged by the tasks they support rather than by whether their coordinates match named variables (Gelada et al. 2019; Huang et al. 2022; K.W. Kim 2026). Here, dynamics-effective is therefore operational rather than semantic. The carrier axes need not correspond to velocity or any other human-interpretable physical coordinate; the intervention can be specified without the test-time native counterfactual hidden state and can launch the intended target-specific future.

## 6.2 From Single control to bounded Joint compositional addressability

The Joint extension shows that the Single-derived interface is not restricted to isolated one-component requests. The same rank-4 carrier and affine map, fitted only on Single examples, can address a bounded same-object request in which both velocity components are changed. This is especially informative for MF1, which receives Single but not Joint counterfactual support during training.

Oracle capacity and addressability remain distinct. Projecting the native Joint hidden difference onto the Single-derived $\mathbf{U}_4$ asks whether that subspace can carry the response when the target hidden change is already known; the Single-only addressable route asks whether a usable patch can be generated from the factual state and Joint request alone. Both are effective on the matched test population, but the result remains bounded: it does not establish arbitrary multi-component or multi-object composition, nor a universal additive control algebra.

## 6.3 Compact intervention entry does not imply low-dimensional dynamical closure

The carrier-relative analyses separate intervention entry from dynamical closure. The MF1-to-MF2 shift toward more additive Joint responses occurs both inside the Single-derived $\mathbf{U}_4$ and in its orthogonal complement. At the same time, composition-defect squared norm is strongly enriched in $\mathbf{U}_4$ relative to its $4/192$ dimensional share. The carrier is therefore composition-relevant, but not the unique locus of composition.

The local recurrent-operator diagnostics give a stronger boundary: only a few percent of the carrier-originating one-step operator strength remains inside $\mathbf{U}_4$, with dominant coupling to the complement. Single and Joint local operators also remain distinguishable after the common rank-4 entry. Thus a low-dimensional intervention interface and a low-dimensional closed state are different objects. The present evidence supports the former, not the latter. Rank 4 should therefore be read only as the smallest passing value on the registered grid for the tested Single procedure. It is not the model's intrinsic state dimension.

### 6.4 Training support is associated with the organization of Joint responses

The MF0/MF1/MF2 comparison shows a family-level association across training regimes, not a simple failure-to-success transition. Native Joint competence is already substantial in MF0 and strong in MF1; MF2 is associated mainly with improved Joint rollout fidelity. The hidden responses show a parallel family-level shift toward lower static composition error from MF0 to MF1 to MF2.

The static composition shift is clearer than the dynamic one, but some MF2 checkpoints still have larger composition errors than some MF1 checkpoints. We therefore interpret it as a family-level shift rather than a consistent improvement at every checkpoint.

### 6.5 Controls and hidden-space geometry are not substitutes for dynamics-effective evidence

The position stress test shows why controls are part of the claim. The intended position patch can satisfy the raw rollout gate, but no-patch and random equal-norm routes do so as well, and wrong-object specificity fails. Raw improvement therefore does not isolate a target-specific dynamics-effective intervention.

The Joint analyses expose a complementary limitation of geometric evidence. The $\mathbf{U}_4$-oracle projection is a useful privileged capacity reference, but capacity does not establish addressability. Likewise, MF2 shows better Joint behavior even though its actual addressable patch is not closer in Euclidean distance to the projected native-Joint oracle than the MF1 patch. Distance to this particular oracle projection is therefore not a sufficient proxy for dynamics-effective control.

Together, these results motivate the same standard: decoder movement, subspace capacity, geometric proximity, and low-dimensional fit are supporting diagnostics. The central evidence is whether a one-shot hidden intervention launches the intended autonomous future and remains distinguishable from plausible alternative perturbations.

### 6.6 Scope, locality, and future directions

The present claims are local. Fresh-checkpoint replication supports recovery of the procedure across independently trained models, but effect strength varies and each checkpoint has its own carrier basis. Temporal tests support recoverability and direct reuse only over the tested $t = 5$–9 window, while the Joint extension is evaluated only at $t = 7$.

The environment is also deliberately controlled: one recurrent architecture, two objects, deterministic dynamics, numeric object-state observations, zero evaluated actions, and a twelve-transition autonomous horizon. The results therefore do not establish the same intervention structure for raw-video world models, stochastic or partially observed dynamics, active control, larger scenes, or real physical systems. It also remains unknown how the required intervention rank changes with hidden width, architecture, object count, or physical complexity.

The most direct next step is to separate compact intervention entry from a fuller effective state. A useful reduced state would need not only a compact intervention interface but also predictive sufficiency and dynamical closure; under partial observability it may additionally need history or belief-like information and may be nonlinear or state dependent. Beyond analysis, these criteria could also become training objectives, allowing future work to test whether more addressable or better-closed latent states improve planning and control.

## 7 Conclusion

We show that a controlled 192-dimensional recurrent world model contains a compact, directly addressable intervention interface for bounded velocity counterfactuals. For Single edits, rank 4 is the smallest tested rank that satisfies the registered development criteria; one patch launches a 12-transition autonomous rollout, the frozen procedure replicates on fresh checkpoints, and the same interface is reusable at nearby anchors. The Single-derived carrier and Single-only affine map also support bounded same-object Joint requests. Across matched training regimes, broader intervention support is associated mainly with better Joint rollout fidelity and more additive native Joint hidden responses.

Mechanistic analyses place a clear boundary on the rank-4 result. Composition-related structure is not confined to the rank-4 subspace, and the recurrent dynamics rapidly couple the injected perturbation to the rest of the hidden state. We therefore interpret rank 4 as a compact intervention-entry interface rather than a closed four-dimensional or intrinsic state. The position-edit stress test further shows that raw outcome improvement without control specificity is insufficient evidence for a dynamics-effective carrier.

Together, these results support an operational view of dynamics-effective latent structure based on addressability, target specificity, and autonomous future propagation. Natural next steps are to test dynamical closure and belief-like state under partial observability and to examine how the interface changes across architectures, hidden widths, action-conditioned settings, and richer physical worlds.

## AI Assistance Disclosure

Generative AI tools, including ChatGPT and OpenAI Codex, were used as research-assistance tools for literature discovery and synthesis, refinement of research questions and experimental protocols, code implementation and debugging, analysis organization, and manuscript preparation. All experimental protocols, scientific decisions, formal runs, interpretation of results, and final claims were reviewed and approved by the authors. AI-generated code, references, analyses, and text were reviewed by the authors and checked against the relevant source code, experimental records, and primary literature where applicable. The authors take full responsibility for the accuracy and integrity of the work.

# Appendix A. Formal Definitions and Notation

This appendix fixes the notation used in Sections 3–4. Environment and model details are in Appendix B; factual/counterfactual construction in Appendix C; carrier and addressability methods in Appendices D–E; evaluation criteria in Appendix F; and the Joint and mechanistic analyses in Appendices H–I.

## A.1 Core notation

Core symbols are grouped below to avoid repetition throughout the appendix.

- $\mathbf{s}_{j,t} \in \mathbb{R}^4$ is object $j$'s primitive state; $\mathbf{s}_t \in \mathbb{R}^8$ is the two-object primitive state; and $\mathbf{o}_t \in \mathbb{R}^{2\times 6}$ is the numeric observation.
- $\mathbf{z}_t \in \mathbb{R}^{192}$ is the full recurrent hidden state manipulated by the intervention analysis. Appendix B denotes the same assembled carrier by $\mathbf{z}_t$.
- $t$ is the anchor index and $H$ the number of autonomous future transitions; the main assay uses $H = 12$.
- $\mathbf{e}_{j,a} \in \mathbb{R}^8$ selects one velocity coordinate in the primitive state, whereas $\mathbf{e} \in \mathbb{R}^4$ is the addressable velocity-request vector used by the coefficient map. Single requests have one nonzero slot; Joint requests have two same-object nonzero slots.
- $\mathbf{z}_t^F$ and $\mathbf{z}_t^{CF}$ are the factual and native-counterfactual anchor states, and $\Delta\mathbf{z}_t^{\text{native}} = \mathbf{z}_t^{CF} - \mathbf{z}_t^F$.
- $\mathbf{U}_r \in \mathbb{R}^{192\times r}$ is the rank-$r$ carrier basis, $\mathbf{c} \in \mathbb{R}^r$ its coefficient vector, and $f_{\text{addr}}$ the addressable map from $(\mathbf{s}_t^F, \mathbf{e})$ to $\mathbf{c}$.
- $R_H$ is the autonomous rollout operator, and $\Omega$ is the bounded state-edit-time domain over which intervention claims are made.

## A.2 Factual state and autonomous rollout

For object $j \in \{A, B\}$, the primitive state is

$$\mathbf{s}_{j,t} = \left[x_{j,t},\, y_{j,t},\, v_{x,j,t},\, v_{y,j,t}\right]^\top \in \mathbb{R}^4 \tag{A1}$$

and the joint two-object state is

$$\mathbf{s}_t = \left[\mathbf{s}_{A,t}^\top,\, \mathbf{s}_{B,t}^\top\right]^\top \in \mathbb{R}^8 \tag{A2}$$

Relative velocity, center velocity, pair distance, and contact are derived quantities rather than additional primitive coordinates. Let $\mathcal{E}$ denote the model history/update map and $R_H$ the autonomous rollout operator:

$$\mathbf{z}_t^F = \mathcal{E}(\mathbf{o}_{0:t}^F) \tag{A3}$$

$$\widehat{\mathbf{Y}}_{t:t+H}^F = R_H(\mathbf{z}_t^F) \tag{A4}$$

Here $\widehat{\mathbf{Y}}_{t:t+H}^F$ is the decoded factual trajectory beginning at the anchor and extending for $H$ autonomous transitions.

### A.3 Same-history local counterfactual

Factual and counterfactual branches share the same pre-anchor observation history:

$$\mathbf{o}_{0:t-1}^{CF} = \mathbf{o}_{0:t-1}^{F} \tag{A5}$$

A Single request changes one velocity coordinate at the anchor:

$$\mathbf{s}_t^{CF} = \mathbf{s}_t^{F} + \delta v\, \mathbf{e}_{j,a}, \qquad a \in \{x, y\} \tag{A6}$$

The selector $\mathbf{e}_{j,a} \in \mathbb{R}^8$ is zero except at the selected velocity coordinate. A Joint request changes both velocity components of the same object:

$$\mathbf{s}_t^{CF} = \mathbf{s}_t^{F} + \delta v_x\, \mathbf{e}_{j,x} + \delta v_y\, \mathbf{e}_{j,y}, \qquad \delta v_x \neq 0,\ \delta v_y \neq 0 \tag{A6a}$$

Positions and the other object's primitive state remain unchanged directly at the anchor. Both Single and Joint edits are restricted to their registered support.

Passing the edited anchor observation through the normal model update gives the native counterfactual reference:

$$\mathbf{z}_t^{CF} = \mathcal{E}(\mathbf{o}_{0:t-1}^{F}, \mathbf{o}_t^{CF}) \tag{A7}$$

$$\Delta \mathbf{z}_t^{\text{native}} = \mathbf{z}_t^{CF} - \mathbf{z}_t^{F} \tag{A8}$$

The same definitions apply to Single and Joint requests; only the edited anchor observation differs.

### A.4 Low-rank and addressable intervention

A rank-$r$ candidate restricts the one-shot hidden perturbation to the column space of $\mathbf{U}_r$:

$$\delta \mathbf{z} = \mathbf{U}_r \mathbf{c} \tag{A9}$$

Addressability requires the coefficients to be predicted without using the evaluation unit's native counterfactual hidden state:

$$\mathbf{c} = f_{\text{addr}}(\mathbf{s}_t^{F}, \mathbf{e}) \tag{A10}$$

$$\mathbf{z}_t^{*} = \mathbf{z}_t^{F} + \mathbf{U}_r \mathbf{c} \tag{A11}$$

The basis orientation is not unique. For any orthogonal $\mathbf{O} \in \mathbb{R}^{r\times r}$,

$$\mathbf{U}_r \mathbf{c} = (\mathbf{U}_r \mathbf{O})(\mathbf{O}^{\top} \mathbf{c}), \qquad \mathbf{O}^{\top}\mathbf{O} = \mathbf{I}_r \tag{A12}$$

Thus individual carrier axes are not interpreted as named physical variables. In the Joint evaluation, $\mathbf{U}_4$ and $f_{\text{addr}}$ remain Single-derived and Single-only fitted. The rollout uses the full affine output $\mathbf{c}$; zero-edit-subtracted variants are geometry diagnostics only (Appendix I).

### A.5 Dynamics-effective criterion and variable roles

The evaluation distinguishes direct targets, quantities derived immediately from the edited anchor state, downstream dynamical consequences, and variables that should remain unaffected. A dynamics-effective intervention must launch the registered counterfactual while satisfying the corresponding preservation, validity, and specificity criteria:

$$\hat{\mathbf{Y}}^{*}_{t:t+H} = R_H(\mathbf{z}^{*}_{t}) \approx \mathbf{Y}^{CF}_{t:t+H} \tag{A13}$$

The symbol ≈ denotes the registered multi-part decision rule, not coordinate-wise equality or a single Euclidean-distance threshold. Exact metrics and controls are defined in Appendix F.

### A.6 Bounded operational domain

Positive intervention claims are restricted to the tested operational domain

$$\Omega = \Omega_{\text{state}} \times \Omega_{\text{int}} \times \Omega_{\text{time}} \tag{A14}$$

Here $\Omega_{\text{state}}$ is the tested factual-state region, $\Omega_{\text{int}}$ the registered edit family and magnitude support, and $\Omega_{\text{time}}$ the evaluated anchor times. The primary Single assay uses $t = 7$, with separate temporal tests over $t \in \{5,6,7,8,9\}$; the Joint extension is evaluated at $t = 7$ within its matched support. Position edits form a separate diagnostic family.

## Appendix B. Environment, Observation Interface, Model Architecture, and Training

This appendix documents the implemented simulator, observation interface, recurrent model, dataset, and training configuration. We report only quantities represented in the scientific interface; unrepresented parameters, such as explicit mass and restitution, are not assigned values.

### B.1 Simulator and physical state

Table B1. Simulator and physical-state specification

| Item | Implemented value |
|---|---|
| Scene | Two circular objects moving in a two-dimensional square reflecting arena |
| Primitive state per object | x, y, horizontal velocity, vertical velocity |
| Two-object primitive state | 2 × 4 array; 8 scalar values |
| Time step | 0.25 |
| Object radius | 0.25 |
| Valid object-center coordinate range | [-4.75, 4.75] on each axis |
| Process noise | None; simulator is deterministic |
| Stored episode | 22 physical states (indices 0–21) and 21 action slots |
| Explicit mass / restitution | Not represented as simulator parameters |

Within each time step, the simulator solves for the earliest circle–circle contact time by setting the distance between the two linearly moving centers equal to the sum of the radii, which gives a quadratic equation in time. It advances to contact, applies the implemented normal impulse update, and then advances the remaining time. Boundary violations are handled by reflection. The implementation is consistent with equal-mass frictionless normal collision behavior, but mass and restitution are not explicit simulator variables.

### B.2 Observation interface

Table B2. Observation interface

| Property | Value |
|---|---|
| Per-time observation shape | 2 × 6 |
| Per-object channels | normalized noisy x, y, horizontal velocity, vertical velocity, visibility, identity |
| Identity values | -1 for one object slot and +1 for the other |
| Directly observed when visible | position, velocity, identity |
| Not directly observed | radius, contact |
| Position-noise standard deviation | 0.005 |
| Velocity-noise standard deviation | 0.01 |
| Visibility schedule | frames 0–4 and the selected anchor are visible; physical channels at other frames are zeroed |

Paired factual and counterfactual branches share the same realized observation-noise array, so carrier differences are not driven by independent noise draws. When visible, position and velocity are provided directly; the study therefore does not claim visual or finite-difference inference of velocity.

### B.3 Recurrent world model

The deterministic belief-like recurrent model has 496200 trainable parameters and exposes a 192-D carrier at the intervention interface. Table B3 lists the carrier blocks and implemented module dimensions; Figures B1–B3 show input encoding, the deterministic recurrent/prior update, and posterior assimilation with final carrier assembly.

Table B3. Recurrent carrier composition and model-module dimensions

| Carrier block | Width |
|---|---|
| Object 0 deterministic recurrent state | 48 |
| Object 0 middle component | 16 |
| Object 1 deterministic recurrent state | 48 |
| Object 1 middle component | 16 |
| Global deterministic recurrent state | 64 |
| Total | 192 |
| **Module** | **Implemented dimensions** |
| Observation encoder | 6 → 128 → 64 |
| Action encoder | 4 → 128 → 64 |
| Message network | 192 → 256 → 128 → 64 |
| Shared object GRU | input 144, hidden 48 |
| Global GRU | input 192, hidden 64 |
| Prior network | 176 → 256 → 128 → 16 |
| Posterior network | 176 → 256 → 128 → 16 |
| Decoder | 192 → 384 → 192 → 8 |
| Normalization layers | none |

### B.3.1 Observation and action encoding

The same 6 → 128 → 64 MLP encodes both 6-D object observations, while visibility is retained separately for prior/posterior selection. The 4-D action is encoded into three 64-D embeddings for the two object updates and the global update. Since no action is involved in all the experiments, all reported experiments use zero actions.

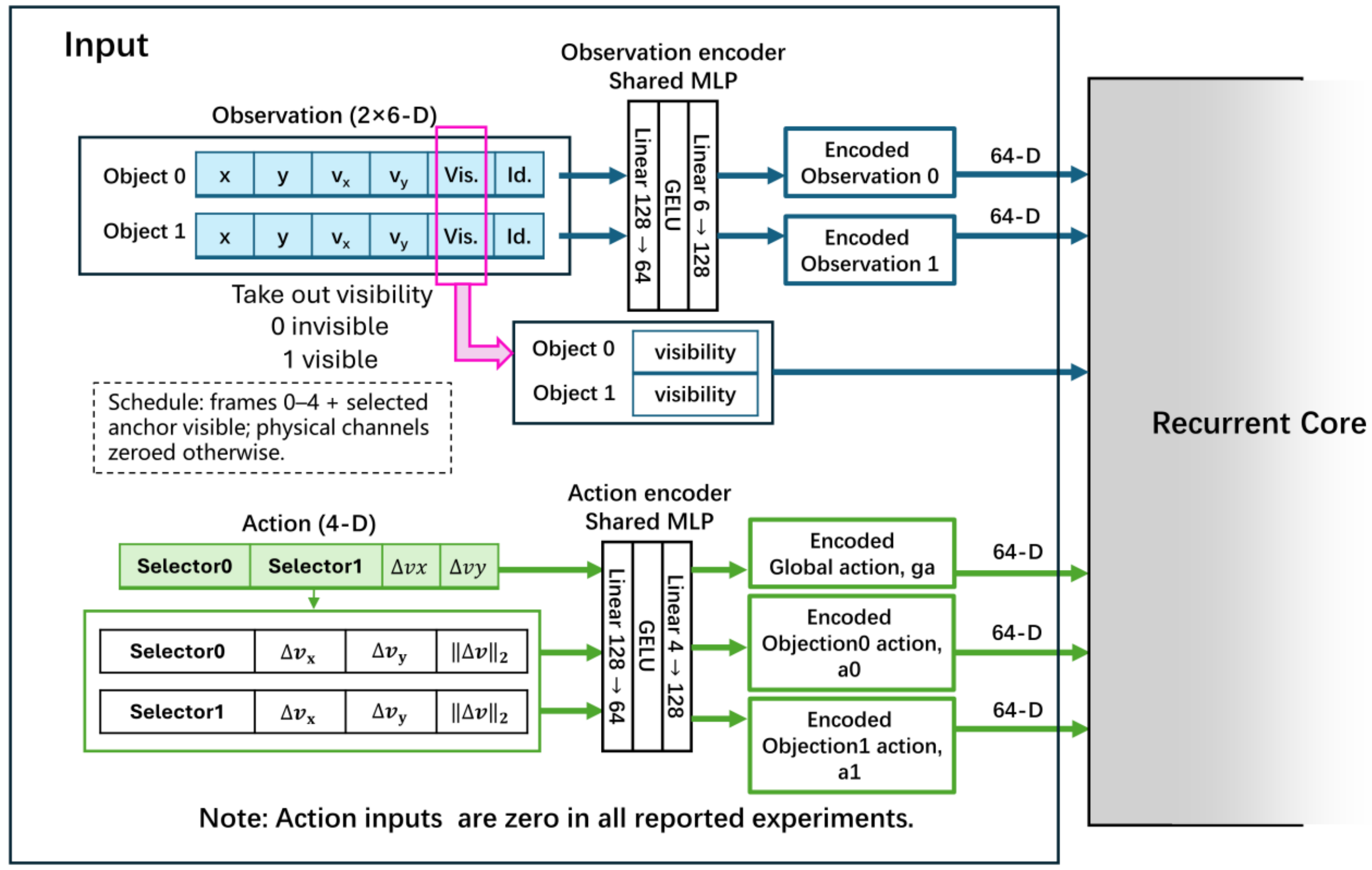

Figure B1. Observation and action encoding. The shared observation encoder produces one 64-D embedding per object and retains visibility separately; the action encoder produces two object-specific and one global 64-D embedding. All reported actions are zero.

### B.3.2 Deterministic recurrent and prior update

The previous carrier is unpacked and the shared message MLP is applied to the ordered and object-swapped object states. For object $i$, the Object GRUCell receives $[\boldsymbol{m}_{\mathrm{i}}, \boldsymbol{message}_{\mathrm{i}}, \boldsymbol{a}_{\mathrm{i}}]$ as a 144-D data input and $\boldsymbol{h}_{\mathrm{i}}$ as a separate 48-D recurrent hidden state, producing $\boldsymbol{h'}_{\mathrm{i}}$. The Global GRU receives $[\boldsymbol{h'}_0, \boldsymbol{m}_0, \boldsymbol{h'}_1, \boldsymbol{m}_1, \boldsymbol{ga}]$ as its 192-D data input and $\boldsymbol{gs}$ as its 64-D recurrent hidden state, producing $\boldsymbol{gs'}$. The shared prior MLP maps $[\boldsymbol{h'}_{\mathrm{i}}, \boldsymbol{gs'}, \boldsymbol{a}_{\mathrm{i}}]$ to a capped 16-D prior middle component.

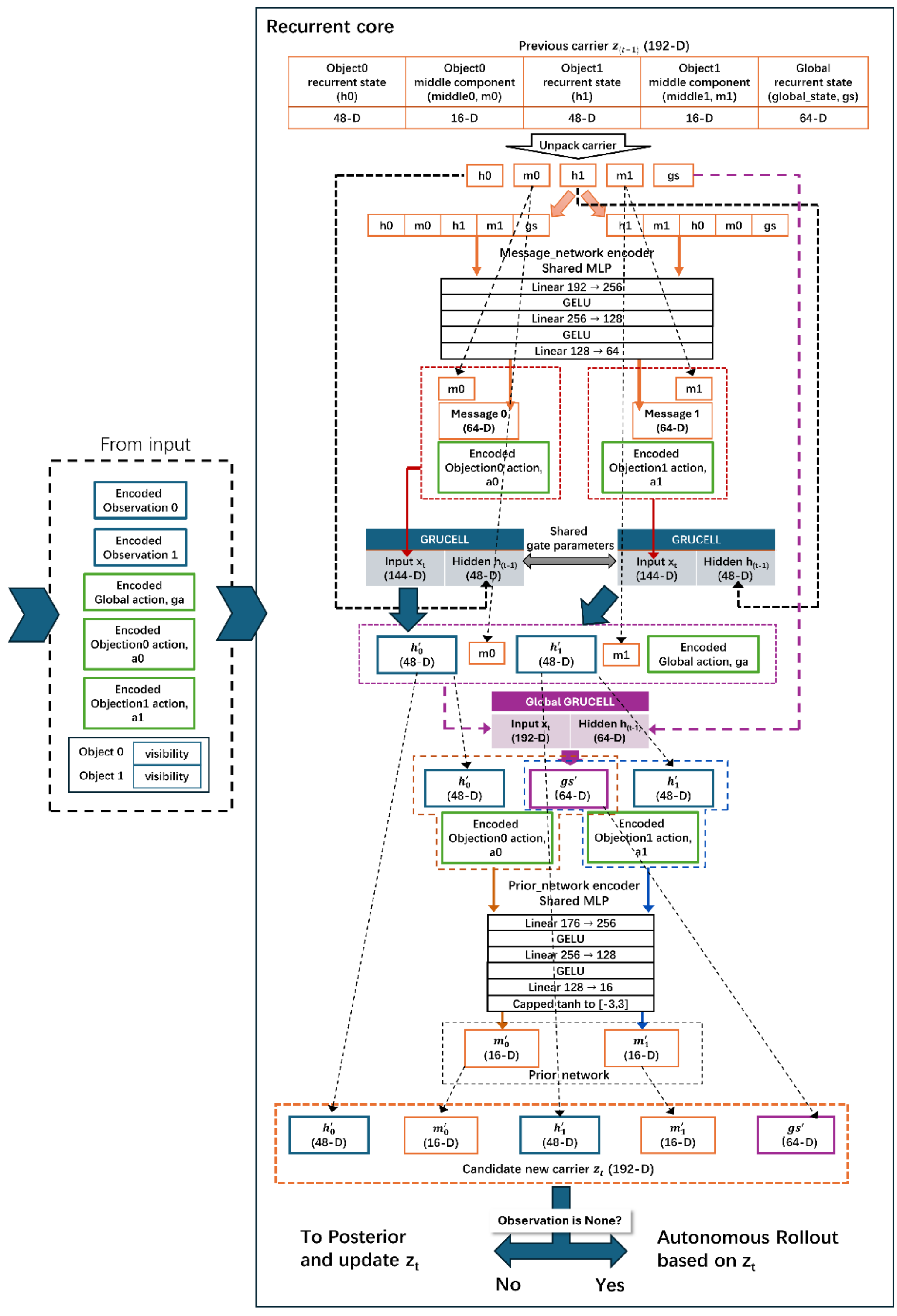

Figure B2. Deterministic recurrent and prior update. The message MLP and Object GRUCells are shared across object slots; the Global GRU produces $\boldsymbol{gs'}$, and the shared prior MLP produces the two 16-D prior middle components.

### B.3.3 Posterior assimilation and carrier assembly

When an anchor observation is available, the shared posterior MLP maps $[\boldsymbol{h}'_{\mathrm{i}}, \boldsymbol{gs}', \boldsymbol{e_obs}_{\mathrm{i}}]$ to a 16-D posterior middle component. With visibility $vis_{\mathrm{i}} \in \{0,1\}$, the selected middle component is

$$\boldsymbol{m}'_{\mathrm{i}} = vis_{\mathrm{i}}\, \boldsymbol{m}^{\mathrm{post}}{}_{\mathrm{i}} + (1 - vis_{\mathrm{i}})\, \boldsymbol{m}^{\mathrm{prior}}{}_{\mathrm{i}}$$

The selected components are packed in fixed order as $\boldsymbol{z}_{\mathrm{t}} = [\boldsymbol{h}'_0, \boldsymbol{m}'_0, \boldsymbol{h}'_1, \boldsymbol{m}'_1, \boldsymbol{gs}'] \in \mathbb{R}^{192}$. The update is deterministic; there is no sampling or reparameterization step.

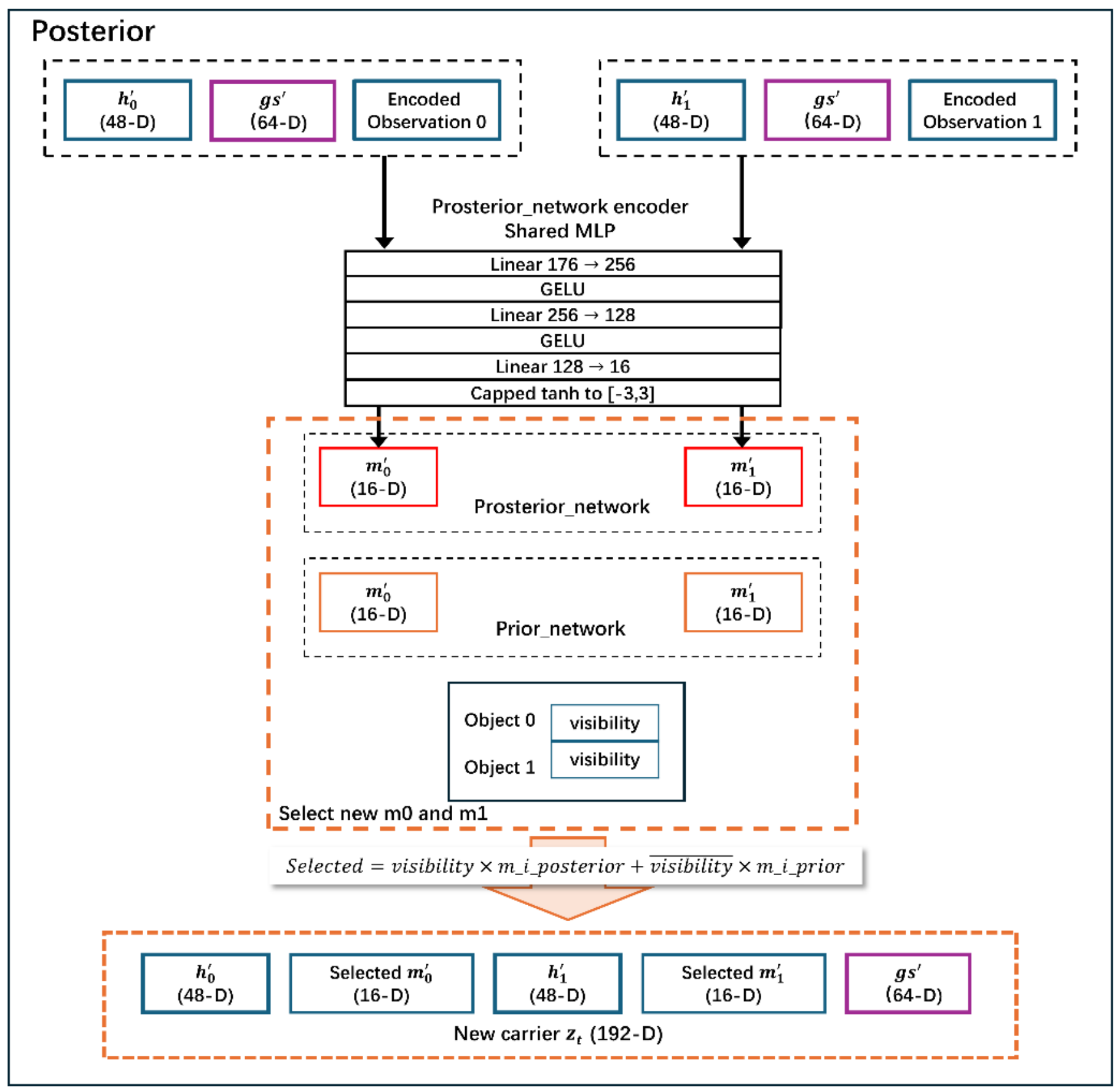

Figure B3. Posterior assimilation and carrier assembly. Visibility selects between prior and posterior middle components before the updated object states, selected middle components, and updated global state are packed in the fixed 48 + 16 + 48 + 16 + 64 layout.

At anchor time $t$, the stored carrier is post-recurrent-update and post-observation-assimilation. The one-shot intervention is applied to this complete 192-D carrier before the future autonomous rollout.

## B.4 Training data and optimization

The dataset was generated specifically for this study with the deterministic simulator in Appendix B.1; no external physics dataset or pretrained trajectory corpus is used. It contains 5120 units, each with a factual trajectory and a same-history local-counterfactual branch sharing the pre-anchor physical history and realized observation noise. At the selected anchor, the registered velocity edit is applied and the simulator is propagated forward from the edited state. Dataset generation used seed 290413 and the data split used seed 290414.

Table B4. Training data and optimization settings

| Training item | Value |
|---|---|
| Dataset units | 5120 total units; 10240 factual/counterfactual branches |
| Training split | 4096 units |
| Development-validation split | 512 units |
| Model-adequacy nonfinal split | 512 units |
| Training anchors | t = 5, 6, 7, 8, 9 |
| Batch construction | 96 units / 192 branches; factual weight 0.5, counterfactual weight 0.5 |
| Training duration | 70 epoch-equivalents; 24 batches per epoch; 1,680 optimizer updates |
| Optimizer | AdamW; learning rate 0.0015; weight decay 0.00001; gradient clip norm 5.0 |
| Scheduler | Cosine annealing over 70 epoch-equivalents; minimum learning rate 0.00015 |
| Checkpoint selection | terminal model state after 1,680 updates; no best-validation selection loop |

Training combines rollout MSE with anchor-state, filtering-state, relative-state, and delta-velocity losses, weighted 1.0, 0.2, 0.2, 0.25, and 0.25. The delta-velocity term uses the recorded q90 scale 1.2762992187. Contact/event labels, carrier/rank supervision, and intervention-method hyperparameter search are not used.

## B.5 Checkpoint roles

Table B5. Development and fresh-replication checkpoint roles

| Public role | Internal seed | Use in this paper |
|---|---|---|
| Development model 1 | 291101 | method development and rank selection |
| Development model 2 | 291102 | method development and rank selection |
| Development model 3 | 291103 | method development and rank selection |
| Fresh replication model 1 | 291301 | frozen-method replication |
| Fresh replication model 2 | 291302 | frozen-method replication |
| Fresh replication model 3 | 291303 | frozen-method replication |

All six checkpoints share the same architecture and 192-D carrier interface. Carrier bases and addressable maps are checkpoint-specific, and numerical basis coordinates are not assumed to align across independently trained checkpoints.

## B.6 Matched training regimes for the compositional extension

For the Joint and mechanistic extension, MF0–MF2 keep the architecture, optimizer, scheduler, update count, and total branch presentations fixed; only the intervention-support composition changes. Single and Joint are support types, not different architectures.

Table B6. Matched training regimes used in the compositional extension

| Family | Per-update branch composition | Role in the paper | Evidence role |
|---|---|---|---|
| MF0: factual-only | 192 factual branches | Contextual comparison for native Joint behavior and hidden-response organization. | Contextual model-family comparison |
| MF1: factual + Single-CF support | 96 factual + 96 Single-counterfactual branches | Primary Single-support family; also serves as the baseline for the compositional extension. | Primary confirmatory family + extension baseline |
| MF2: factual + Single+Joint-CF support | 96 factual + 48 Single-counterfactual + 48 Joint-counterfactual branches | Tests how explicit Joint support is associated with Joint rollout fidelity, specificity, and hidden-response organization. | Replicated Joint-support extension |

Each family uses 1680 updates and 192 branch presentations per update (322560 total). In MF2, a Joint-counterfactual branch changes both velocity components of one object at the anchor while leaving all positions and the other object's primitive state unchanged directly. Appendix C defines the exact Joint semantics. These extension families do not alter the original rank-selection procedure or its confirmatory status.

### B.7 Dataset and analysis split summary

Table B7. Dataset and analysis split

| Purpose | Units | Where used |
|---|---|---|
| Model-training dataset | 5120 total | B.4 |
| Training split | 4096 | model fitting |
| Development-validation | 512 | model development |
| Model-adequacy nonfinal | 512 | adequacy check |
| Single carrier-fit | 1024 / checkpoint | carrier + affine map fitting |
| Main Single confirmatory | 256 / checkpoint / stratum | final Single assay |
| Temporal B1 carrier fit | 128 / anchor / stratum | temporal fit |
| Temporal B1 method dev | 128 / anchor / stratum | temporal development |
| Temporal B1 confirmatory | 128 / anchor / stratum | temporal evaluation |
| Joint FIT | 411 / checkpoint | whitening/operator fit |
| Joint TEST | 101 / checkpoint | composition/addressability/specificity |

## Appendix C. Counterfactual Pair Construction and Intervention Semantics

This appendix fixes the physical counterfactual semantics used by the Single, Joint, and position-edit assays: pairing, anchor edits, simulator truth, reference routes, strata, and support.

### C.1 Same-history pairing

Each factual/counterfactual pair shares the same pre-anchor physical trajectory and the same realized observation-noise sequence:

$$\mathbf{s}_{0:t-1}^{\mathrm{CF}} = \mathbf{s}_{0:t-1}^{\mathrm{F}} \qquad \boldsymbol{\epsilon}_{0:t}^{\mathrm{CF}} = \boldsymbol{\epsilon}_{0:t}^{\mathrm{F}} \tag{C1}$$

Here $\mathbf{s}$ denotes the primitive physical state and $\boldsymbol{\epsilon}$ the realized observation noise. The counterfactual branch first differs at the registered anchor edit. Joint units obey the same pairing; their identity contract is specified in Section C.7.

### C.2 Primary Single velocity-edit family

The velocity request is represented by a four-slot vector **e** ordered as $\left(\Delta v_{x,A}, \Delta v_{y,A}, \Delta v_{x,B}, \Delta v_{y,B}\right)$. A Single request has exactly one nonzero slot:

$$\mathbf{e} = [\Delta v_{x,A} \quad \Delta v_{y,A} \quad \Delta v_{x,B} \quad \Delta v_{y,B}]^T \qquad \| \mathbf{e} \|_0 = 1 \tag{C2}$$

The registered Single family uses the following signed increments:

$$\delta v_{\mathrm{abs}} \in \{0.066,\ 0.132,\ 0.264\} \qquad \delta v \in \{-\delta v_{\mathrm{abs}}, +\delta v_{\mathrm{abs}}\} \tag{C3}$$

At the anchor, all positions and the other three velocity coordinates remain unchanged. Derived quantities may change immediately; later positions, contact, and velocities arise only through simulator evolution. The edited-anchor speed remains within the registered support (maximum recorded bound approximately 0.6711).

Joint uses the same four-slot request but changes both velocity coordinates of one object:

$$\| \mathbf{e} \|_0 = 2 \qquad \mathrm{supp}(\mathbf{e}) = \{(j,x),(j,y)\} \qquad j \in \{A,B\} \tag{C2a}$$

The two-component Euclidean request magnitude and edited-anchor speed remain within the registered Single support. All other primitive anchor coordinates remain unchanged directly.

### C.3 Simulator truth and model-native counterfactual

After the anchor edit, the deterministic simulator is propagated from the edited state to define counterfactual truth. G1 passes the edited anchor observation through the model's ordinary update and then releases the model without future counterfactual observations. G2 instead adds the complete native hidden difference to the factual hidden state:

$$\mathbf{z}_t^{G2} = \mathbf{z}_t^{\mathrm{F}} + \left(\mathbf{z}_t^{\mathrm{CF}} - \mathbf{z}_t^{\mathrm{F}}\right) = \mathbf{z}_t^{\mathrm{CF}}. \tag{C4}$$

G1 against simulator truth is the model-adequacy test; G2 is only an interface-equivalence control. The same definitions apply to Joint units, with the base identity and requested velocity vector fixed before evaluation. Neither route establishes low-rank capacity or addressability.

### C.4 Evaluation strata

Table C1. Evaluation strata and their operational roles

| Stratum | Operational role |
|---|---|
| S1: NO_CONTACT_WITHIN_HORIZON | Neither the factual nor the counterfactual trajectory contains a detected future contact step over the 12-transition horizon; used to test local kinematics and autonomous propagation without later contact. |
| S2: EDIT_AFFECTS_FUTURE_CONTACT | The factual trajectory contains a detected future contact step and the edit produces a sufficiently large post-anchor factual–counterfactual trajectory difference; used to test downstream coupled consequences of the edit. Counterfactual contact, a contact/noncontact switch, or a changed contact time is not required. |

S1 and S2 are scored separately; success in S1 cannot compensate for failure in S2.

### C.5 Anchor and rollout support

The main Single assay uses anchor $t = 7$ and $H = 12$ autonomous transitions. Temporal tests use $t \in \{5,6,7,8,9\}$, including transport of the exact $t = 7$ interface. Joint is evaluated at $t = 7$ with the same horizon. Because $t$ is a discrete step index (simulator step 0.25), the claims remain restricted to the registered edit and edited-anchor speed support.

### C.6 Position-edit diagnostic boundary

The position stress test is development-only. At $t = 7$, one $x$ or $y$ coordinate of one object is shifted by 0.05, 0.10, or 0.20; the visible normalized anchor observation is updated consistently, while all velocities and other primitive anchor coordinates remain unchanged. Simulator truth is regenerated from the edited state. The intended patch is compared with no-patch, random equal-norm, and wrong-object alternatives. This diagnostic neither selects the velocity-carrier rank nor enters fresh-checkpoint confirmation.

### C.7 Same-object two-component Joint velocity extension

For each Joint request at t=7, we first identify the corresponding factual unit, then verify the edited object and velocity change directly from the factual-to-Joint anchor difference. Composition uses the matched quadruplet $(F_i, S_{x,i}, S_{y,i}, J_{xy,i})$: all four routes share the base unit, edited object, anchor, pre-anchor history, and realized observation noise, and the two Single request vectors sum exactly to the Joint request. The 512 base units are split before analysis into 411 FIT and 101 TEST units (Appendix H.1).

Joint addressability still uses the interface Single-derived. For MF1 and MF2, $\mathbf{U}_4$ and $f_{\mathrm{addr}}$ are fitted checkpoint-specifically from Single examples only; no Joint hidden difference or Joint map example is used for fitting. The actual Joint rollout uses the full affine prediction from the factual state and Joint request. Thus MF1 tests unseen-Joint generalization, whereas MF2 tests the same addressability construction in a model trained with Joint support. Appendix H defines the held-out route comparison and matched controls. The extension does not reopen Single rank selection.

# Appendix D. Low-Rank Carrier Construction

This appendix specifies the checkpoint-specific linear carrier used by the intervention assay. The basis is fitted only from Single carrier-fit hidden differences. Joint hidden differences never enter the basis fit; in the later extension they are used only for privileged projection and geometry diagnostics.

## D.1 Hidden-difference matrix

For Single carrier-fit unit $i$ at anchor $t$, define the native counterfactual-minus-factual carrier difference at the same post-observation/pre-transition interface:

$$\Delta \mathbf{z}_i = \mathbf{z}_{t,i}^{CF} - \mathbf{z}_{t,i}^{F} \in \mathbb{R}^{192} \tag{D1}$$

Each checkpoint contributes $N = 1024$ Single fit units, stacked row-wise as

$$\mathbf{D} = \begin{bmatrix} (\Delta \mathbf{z}_1)^\top \\ \vdots \\ (\Delta \mathbf{z}_N)^\top \end{bmatrix} \in \mathbb{R}^{N\times 192} \qquad N = 1024 \tag{D2}$$

The matrix $\mathbf{D}$ contains raw, uncentered fit-split differences only. Test units and later development or confirmatory outcomes do not enter the basis fit.

## D.2 Uncentered singular-value decomposition

We factor the raw difference matrix without subtracting its mean:

$$\mathbf{D} = \mathbf{L}\,\mathbf{\Sigma}\,\mathbf{V}^\top \tag{D3}$$

Here $\mathbf{L}$ and $\mathbf{V}$ contain the left and right singular vectors, and $\mathbf{\Sigma} = \operatorname{diag}(\sigma_1, \dots, \sigma_{192})$ contains the singular values. If $\mathbf{v}_k$ is the $k$th right singular vector, the candidate rank-$r$ carrier basis is

$$\mathbf{U}_r = [\mathbf{v}_1, \dots, \mathbf{v}_r] \in \mathbb{R}^{192\times r} \qquad \mathbf{U}_r^\top \mathbf{U}_r = \mathbf{I}_r \tag{D4}$$

For deterministic artifact storage, each $\mathbf{v}_k$ is signed so that its largest-magnitude loading is positive. This convention has no scientific meaning because flipping a basis vector and its coefficient leaves the 192-D perturbation unchanged.

## D.3 Projection and oracle capacity

The rank-$r$ projector, privileged oracle coordinates, and projected reconstruction are

$$\mathbf{P}_r = \mathbf{U}_r \mathbf{U}_r^\top$$

$$\mathbf{c}_i^{\text{oracle}} = \mathbf{U}_r^\top \Delta \mathbf{z}_i$$

$$\widehat{\Delta \mathbf{z}}_i^{\text{oracle}} = \mathbf{U}_r \mathbf{c}_i^{\text{oracle}} = \mathbf{P}_r \Delta \mathbf{z}_i \tag{D5}$$

Thus $\widehat{\Delta\mathbf{z}}_i^{\text{oracle}}$ is the closest Euclidean projection of the native hidden difference onto $\text{col}(\mathbf{U}_r)$. Because $\Delta\mathbf{z}_i$ uses the same unit's native counterfactual hidden state, this route measures carrier capacity rather than test-time addressability.

For Joint unit $i$, let $\Delta\mathbf{z}_i^J = \mathbf{z}_{t,i}^J - \mathbf{z}_{t,i}^F$. With the Single-derived $\mathbf{U}_4$ frozen,

$$\mathbf{c}_i^{J,U_4\text{-oracle}} = \mathbf{U}_4^\top \Delta\mathbf{z}_i^J$$

$$\widehat{\Delta\mathbf{z}}_i^{J,U_4} = \mathbf{P}_4 \Delta\mathbf{z}_i^J$$

(D5a)

These quantities are privileged Joint-capacity references only; they are not used to generate the primary addressable Joint patch.

### D.4 Descriptive second-moment capture

As a descriptive geometry summary, we report the cumulative squared-singular-value fraction

$$\rho_r = \frac{\sum_{k=1}^{r} \sigma_k^2}{\sum_{k=1}^{192} \sigma_k^2}$$

(D6)

Because the SVD is uncentered, $\rho_r$ describes raw second-moment concentration, not mean-centered variance. It is not a physical-energy measure, an intrinsic-dimension estimate, or the rank-selection criterion.

### D.5 Registered rank sets

$$\mathcal{R} = \{1,2,4,8,12,16,20,32\} \qquad \mathcal{R}_{\text{diag}} = \{64\} \qquad \mathcal{R}_{\text{ref}} = \{0,192\}$$

(D7)

The selectable grid $\mathcal{R}$ is fixed before development outcomes are read. Rank 64 is diagnostic; ranks 0 and 192 are no-patch and full-hidden references. The reported rank is therefore the smallest passing tested rank on $\mathcal{R}$, not a claim about untested ranks or nonlinear carrier families.

### D.6 Checkpoint and split boundaries

Each checkpoint fits its own basis from its 1024-unit carrier-fit split; no cross-checkpoint axis alignment is assumed. Method-development data may evaluate that basis but do not refit it, and confirmatory data do not enter rank selection or method changes. Fresh checkpoints use the already frozen construction.

The Joint extension does not reopen the rank grid or selection rule; it reuses the rank selected by the primary Single development panel.

### D.7 Single-derived rank-4 carrier in the Joint extension

For MF1 and MF2, $\mathbf{U}_4$ remains checkpoint-specific and is fitted from Single hidden differences only; Joint hidden differences are excluded from basis fitting.

The frozen $\mathbf{U}_4$ then has two roles: Eq. (D5a) gives a privileged Joint-capacity projection, whereas Appendix E uses the same Single-derived basis with the Single-only addressable map $f_{\text{addr}}$ to generate the actual Joint patch. Joint hidden differences are used for the former, not to fit the latter.

MF0 enters native Joint behavior and composition comparisons but not the Single-derived $\mathbf{U}_4$ addressability analysis. The factual-whitened projectors introduced in Appendix I are diagnostic coordinate transformations only; they neither refit $\mathbf{U}_4$ nor change the rollout patch.

## Appendix E. Addressable Coefficient Map

The carrier basis specifies a candidate low-dimensional subspace; the addressable map specifies how to reach it from a factual physical state and requested edit. The map is fitted only on Single carrier-fit examples and is reused unchanged for the Joint test. At evaluation time it never reads the unit's native counterfactual hidden state.

### E.1 Input encoding

$$\mathbf{x}_i = \begin{bmatrix} \mathbf{e}_i \\ \mathbf{s}_{t,i}^F \end{bmatrix} \in \mathbb{R}^{12}$$
$$\mathbf{e}_i \in \mathbb{R}^4, \qquad \mathbf{s}_{t,i}^F \in \mathbb{R}^8$$

(E1)

The four entries of $\mathbf{e}_i$ are the horizontal and vertical velocity edits for objects A and B. Single fitting uses one nonzero slot; Joint evaluation uses the two same-object velocity slots. The factual state $\mathbf{s}_{t,i}^F$ contributes the eight primitive position and velocity coordinates. No Joint-specific input feature is added.

### E.2 Fit-only standardization

Each non-intercept feature is standardized with carrier-fit statistics only:

$$\tilde{x}_{i,k} = \frac{x_{i,k} - \mu_k}{\max(\sigma_k, 10^{-6})} \qquad k = 1, \dots, 12$$

(E2)

Here $\mu_k$ and $\sigma_k$ are the fit-split mean and standard deviation. The floor $10^{-6}$ avoids division by a negligible scale. These statistics are frozen for evaluation and are not recomputed in the temporal B2 transport test.

### E.3 Coefficient targets

Single carrier-fit units provide privileged coefficient targets by projection onto the checkpoint-specific carrier:

$$\mathbf{c}_i^{\text{oracle}} = \mathbf{U}_r^{\top} \Delta \mathbf{z}_i$$
$$\mathbf{C}_r = \mathbf{D}\mathbf{U}_r \in \mathbb{R}^{N \times r}$$

(E3)

Row $i$ of $\mathbf{C}_r$ is the target coefficient vector for one Single fit unit; $\mathbf{D}$ and $\mathbf{U}_r$ are defined in Appendix D. Joint hidden differences are excluded from the mapper targets.

### E.4 Ridge-affine fit

$$(\mathbf{W}_r, \mathbf{b}_r) = \arg\min_{\mathbf{W},\mathbf{b}} \sum_{i=1}^{N} \left\| \mathbf{c}_i^{\text{oracle}} - \left(\mathbf{W}^{\top} \tilde{\mathbf{x}}_i + \mathbf{b}\right) \right\|_2^2 + \lambda \parallel \mathbf{W} \parallel_F^2$$
$$\lambda = 10^{-4}$$

(E4)

Eq.(E4) fits the parameters $\boldsymbol{W}_r$ and $\boldsymbol{b}_r$ of the rank-specific addressable map $f_{\text{addr},r}$, which is defined explicitly in Eq. (E5). The ridge penalty applies to $\mathbf{W}_r$ but not to the affine intercept $\mathbf{b}_r$. The mapper family, regularization, and fit protocol are fixed from the Single procedure; no Joint example, nonlinear alternative, or outcome-based mapper selection is introduced for the Joint test.

### E.5 Test-time inference and patch

$$\hat{\mathbf{c}} = f_{\text{addr},r}(\mathbf{s}_t^F, \mathbf{e}) = \mathbf{W}_r^{\top} \tilde{\mathbf{x}} + \mathbf{b}_r$$

(E5)

At test time, $f_{\text{addr},r}$ receives only the factual primitive state and requested edit. It cannot access the native counterfactual hidden state, true hidden difference, simulator future, future observations, or oracle coefficients. The predicted coefficient is converted to the one-shot hidden patch:

$$\mathbf{z}_t^* = \mathbf{z}_t^F + \beta \mathbf{U}_r \hat{\mathbf{c}} \qquad \beta = 1$$

(E6)

After the patch is formed, the addressable map is not consulted again during rollout. This separation is the operational meaning of addressability used in the paper.

## E.6 Oracle capacity versus addressability

Table E1. Oracle-capacity and addressable coefficient routes

| Route | Coefficient source | Allowed interpretation |
|---|---|---|
| Oracle capacity | Same-unit native hidden difference projected onto the rank-r carrier basis. | Tests whether the rank-r subspace can carry the required hidden difference; privileged upper bound only. |
| Addressable coefficient map | Predicted from the factual primitive state and the registered edit. | Primary intervention route; tests whether the carrier can be reached without a test-time native counterfactual hidden state. |

## E.7 Single-only affine composition for Joint requests

The Joint request uses the same four-slot representation and the same checkpoint-specific affine map fitted from Single examples only. No Joint example enters standardization, coefficient fitting, regularization, or mapper selection. For a fixed factual state, let $\mathbf{e}_x$ and $\mathbf{e}_y$ be the two corresponding Single edits. The Joint request and its affine coefficient are defined by

$$\mathbf{e}_{xy} = \mathbf{e}_x + \mathbf{e}_y$$
$$\mathbf{c}_x = f_{\text{addr},4}(\mathbf{s}_t^F, \mathbf{e}_x) \qquad \mathbf{c}_y = f_{\text{addr},4}(\mathbf{s}_t^F, \mathbf{e}_y)$$
$$\mathbf{c}_0 = f_{\text{addr},4}(\mathbf{s}_t^F, \mathbf{0})$$
$$\mathbf{c}_{xy} = \mathbf{c}_x + \mathbf{c}_y - \mathbf{c}_0$$

(E7)

Because the standardized map is affine and the factual state is fixed, the direct Joint coefficient and the component-wise affine composition agree up to numerical precision. This identity concerns the mapper only; it does not imply that the model's native Joint hidden response is additive or that the resulting intervention is dynamics-effective.

The actual Joint rollout uses $\mathbf{c}_{\text{full}} = f_{\text{addr},4}(\mathbf{s}_t^F, \mathbf{e}_{xy}) = \mathbf{c}_{xy}$. The quantities $\mathbf{c}_{\text{base}}$ and $\mathbf{c}_{\text{request}}$ are introduced only for the geometry decomposition in Appendix I and never replace the full rollout coefficient. Rollout behavior and matched specificity controls determine whether the generated Joint patch is dynamics-effective.

## Appendix F. Rollout Metrics, Eligibility Gates, and Registered Thresholds

This appendix gives the exact evaluation kernel for the primary Single low-rank and temporal experiments. The Joint extension reuses the same normalization scales, autonomous-rollout structure, physical-validity checks, and metric roles; only the two-direct-target scoring and later specificity summaries are modified in F.10. Scientific metrics (M1–M5), engineering eligibility (M6E), and rollout-structure checks (M7) are kept separate.

### F.1 Evaluation notation and frozen normalization scales

For evaluation unit $i$, let $t_i$ be the intervention anchor and $H = 12$ the number of autonomous future transitions. The denormalized decoded state at relative index $j$ is $\widehat{\mathbf{X}}_{i,j} \in \mathbb{R}^{2\times4}$, with object rows $o \in \{0,1\}$ and columns $(x, y, v_x, v_y)$. The anchor is $j = 0$ and the released future is $j = 1, \dots, H$. The route-specific reference is $\mathbf{X}^{\mathrm{ref}}_{i,j}$: counterfactual truth for patched, native-CF, wrong-object, and G2 routes, and factual truth for the factual/no-edit route.

The edited object is $o_i \in \{0,1\}$ and the velocity-axis selector is $a_i \in \{0,1\}$, so the edited primitive column is $c_i = 2 + a_i$. Primitive errors are normalized coordinatewise by the frozen scale matrix $\mathbf{s}^P$.

Table F1. Frozen primitive-coordinate normalization scales $\mathbf{s}^P$

| | $x$ | $y$ | $v_x$ | $v_y$ |
|---|---|---|---|---|
| Object 0 (A) | 0.64721 | 0.62436 | 0.40208 | 0.39198 |
| Object 1 (B) | 0.68680 | 0.55883 | 0.40071 | 0.35781 |

M2 uses the 13-dimensional derived representation $\boldsymbol{\phi}(\mathbf{X})$ listed in Table F2 and its frozen scale vector $\mathbf{s}^D$.

Table F2. Frozen scales for the 13-dimensional derived representation

| d | Derived feature $\phi_d$ | Frozen scale $s_d^D$ |
|---|---|---|
| 1 | relative_position_x | 1.04125 |
| 2 | relative_position_y | 0.87639 |
| 3 | relative_velocity_x | 0.74275 |
| 4 | relative_velocity_y | 0.79797 |
| 5 | center_position_x | 0.32511 |
| 6 | center_position_y | 0.32665 |
| 7 | center_velocity_x | 0.06288 |
| 8 | center_velocity_y | 0.07177 |
| 9 | summed_velocity_x | 0.12576 |
| 10 | summed_velocity_y | 0.14354 |
| 11 | 0.5 x sum of squared speeds | 0.06497 |
| 12 | inter-object distance | 0.35839 |
| 13 | $\mathbf{1}\{\text{distance} \le 0.50000001\}$ | 1.00000 |

The scale registry was estimated once from 128 non-final threshold-calibration counterfactual simulator-truth units generated under the same registered Single-edit simulator protocol described in Appendix C. These units followed the frozen intervention-support and feasibility rules and were not selected using model, low-rank, or final-test outcomes. The resulting scales were then frozen for all subsequent evaluations.

For continuous coordinates, the stored scale estimator is $\max(\mathrm{IQR}/1.349,\ 1.4826\,\mathrm{MAD},\ 0.01)$; the contact-indicator scale is 1. These frozen scales are reused by all evaluators. M1–M4 add no clipping, epsilon, intervention-amplitude normalization, or metric-specific weighting.

## F.2 Anchor fidelity and future rollout metrics (M1-M4)

M1 is the normalized absolute error of the directly edited velocity coordinate at the decoded anchor:

$$M_{1,i} = \frac{\left|\widehat{\mathbf{X}}_{i,0,o_i,c_i} - \mathbf{X}^{\mathrm{ref}}_{i,0,o_i,c_i}\right|}{s^{P}_{o_i,c_i}} \qquad c_i = 2 + a_i$$

(F1)

M2 is the unweighted RMS error of the 13 normalized derived quantities at the anchor:

$$M_{2,i} = \sqrt{\frac{1}{13}\sum_{d=1}^{13}\left(\frac{\phi_d\big(\widehat{\mathbf{X}}_{i,0}\big) - \phi_d\big(\mathbf{X}^{\mathrm{ref}}_{i,0}\big)}{s^{D}_{d}}\right)^2}$$

(F2)

M3 is the normalized RMSE over all 12 released future states, both objects, and all four primitive fields; the anchor is excluded:

$$M_{3,i} = \sqrt{\frac{1}{8H}\sum_{j=1}^{H}\sum_{o=0}^{1}\sum_{c=0}^{3}\left(\frac{\widehat{\mathbf{X}}_{i,j,o,c} - \mathbf{X}^{\mathrm{ref}}_{i,j,o,c}}{s^{P}_{o,c}}\right)^2} \qquad H = 12$$

(F3)

M4 is the maximum normalized anchor error over primitive coordinates declared unaffected by the edit:

$$M_{4,i} = \max_{(o,c):\,U_{i,o,c}=1} \frac{\left|\widehat{\mathbf{X}}_{i,0,o,c} - \mathbf{X}^{\mathrm{ref}}_{i,0,o,c}\right|}{s^{P}_{o,c}}$$

(F4)

Here $\mathbf{U}_i \in \{0,1\}^{2\times 4}$ is the unaffected-coordinate mask. Single edits leave seven primitive slots unaffected. Joint direct-target scoring is defined separately in F.10.

## F.3 Post-edit law and geometry consistency (M5)

M5 is the maximum of five checks: one-step simulator replay, circle overlap, arena-boundary violation, excessive per-step displacement, and agreement between the recorded anchor difference and the declared edit. Let $\mathcal{S}$ be the exact one-step simulator used by the evaluator.

$$R^{\mathrm{dyn}}_{i,j} = \sqrt{\frac{1}{8}\sum_{o=0}^{1}\sum_{c=0}^{3}\left(\frac{\widehat{\mathbf{X}}_{i,j+1,o,c} - \left[\mathcal{S}\big(\widehat{\mathbf{X}}_{i,j}, \mathbf{a}_{i,t_i+j}\big)\right]_{1,o,c}}{s^{P}_{o,c}}\right)^2} \qquad 0 \le j < H$$

(F5)

$$R^{\mathrm{overlap}}_{i} = \frac{1}{0.01}\max_{0\le j\le H} \max\big(0.5 - \|\,\widehat{\mathbf{p}}_{i,j,1} - \widehat{\mathbf{p}}_{i,j,0}\,\|_2,\ 0\big)$$

(F6)

$$R^{\mathrm{arena}}_{i} = \frac{1}{0.01}\max_{j,o,q\in\{x,y\}} \max\big(\left|\hat{p}_{i,j,o,q}\right| - 4.75,\ 0\big)$$

(F7)

$$R_i^{\text{step}} = \frac{1}{0.01} \max_{\substack{0\le j<H \\ o\in\{0,1\}}} \max\left(\| \hat{\mathbf{p}}_{i,j+1,o} - \hat{\mathbf{p}}_{i,j,o} \|_2 - 0.25\max\left(\| \hat{\mathbf{v}}_{i,j,o} \|_2, \| \hat{\mathbf{v}}_{i,j+1,o} \|_2\right) - 10^{-8}, 0\right)$$

(F8)

$$\mathbf{B}_i = \mathbf{X}_{i,0}^{CF} - \mathbf{X}_{i,0}^{F} \qquad R_i^{\text{decl}} = \frac{\| \mathbf{B}_i - \mathbf{D}_i \|_\infty}{10^{-8}}$$

(F9)

Here $\mathbf{D}_i \in \mathbb{R}^{2\times 4}$ is zero except at the declared edited slot, where it contains the recorded factual-to-counterfactual anchor change. This term audits the intervention record rather than the model decode.

$$M_{5,i} = \max\left\{\max_{0\le j<H} R_{i,j}^{\text{dyn}}, R_i^{\text{overlap}}, R_i^{\text{arena}}, R_i^{\text{step}}, R_i^{\text{decl}}\right\}$$

(F10)

The terms are not summed or weighted. Collision and reflecting-wall consistency enter through the exact replay. Nonfinite decoded outputs make the law-residual path fail.

## F.4 M6E numerical eligibility

M6E is an engineering eligibility gate, not a scientific metric. The main evaluation requires finite final hidden state and decoded trajectory together with loose explosion bounds:

$$E_i^{\text{main}} \Leftrightarrow \text{finite}\left(\mathbf{z}_i^{\text{final}}\right) \wedge \text{finite}\left(\hat{\mathbf{X}}_{i,0:H}\right) \wedge \| \mathbf{z}_i^{\text{final}} \|_2 \le 10^6 \wedge \| \hat{\mathbf{X}}_{i,0:H} \|_\infty \le 10^6$$

(F11)

The temporal evaluation adds a per-decoded-state vector bound:

$$E_i^{\text{temp}} \Leftrightarrow E_i^{\text{main}} \wedge \max_{0\le j\le H} \| \text{vec}\left(\hat{\mathbf{X}}_{i,j}\right) \|_2 \le 10^6$$

(F12)

M6E does not separately check intermediate hidden states or general execution failures. Failed or ineligible units still count in the denominator, and the same gate is used for Joint reference, oracle, and addressable rollouts.

## F.5 M7 autonomous-rollout structure

M7 records the required structure of every scored 12-transition rollout.

Table F3. Required structural metadata for the autonomous rollout (M7)

| M7 field | Required value |
|---|---|
| transition_count | 12 |
| decoder_count | 13 |
| observation_reads_after_anchor | 0 |
| teacher_forcing_steps | 0 |
| state_clamp_steps | 0 |
| future_oracle_reads | 0 |
| pass | True |

The 13 decodes comprise the anchor plus twelve future states. These structural requirements are unchanged for Joint requests: only the one-shot anchor patch differs.

### F.6 G2 full-hidden interface-equivalence check

G2 tests the patching interface by starting from a copy of the same native counterfactual hidden state used by G1. The decoded-rollout discrepancy is

$$d_Y = \max_{i,j,o,c} \left| Y^{G2}_{i,j,o,c} - Y^{G1}_{i,j,o,c} \right| \qquad \text{G2Pass} \Leftrightarrow d_Y \le 10^{-6} \tag{F13}$$

The hidden discrepancy is zero by construction because both routes start from the same $\mathbf{z}^{CF}$. G2 is therefore an interface-equivalence control, not evidence for low-rank capacity or addressability.

### F.7 Programmatic S1/S2 strata

Define the future contact-step detector $\Gamma$ for an anchor-plus-future state sequence $\mathbf{X}$, where $\mathbf{V}_j \in \mathbb{R}^{2\times2}$ contains both object velocities:

$$\Gamma(\mathbf{X}) = \left\{ j \in \{1, \dots, H\} : \| \mathbf{p}_{j,1} - \mathbf{p}_{j,0} \|_2 \le 0.500001 \ \vee \| \mathbf{V}_j - \mathbf{V}_{j-1} \|_F > 10^{-5} \right\} \tag{F14}$$

The anchor $j = 0$ is excluded from $\Gamma$.

$$S1 \Leftrightarrow \Gamma(\mathbf{X}^F_{0:H}) = \varnothing \ \wedge \ \Gamma\left(\mathbf{X}^{CF}_{0:H}\right) = \varnothing \tag{F15}$$

$$S2 \Leftrightarrow \Gamma(\mathbf{X}^F_{0:H}) \neq \varnothing \ \wedge \ \max_{k > t_i, o, c} \left| X^{CF}_{k,o,c} - X^F_{k,o,c} \right| \ge 10^{-4} \tag{F16}$$

S1 isolates trajectories for which neither branch contains a detected contact step over the 12-transition evaluation window. The detector is applied only to future indices $j = 1, \dots, H$, so the anchor itself is not classified as a contact step. A future step is flagged when either the two object centers come within the registered contact-distance tolerance or the joint velocity array exhibits a change larger than $10^{-5}$. Thus, S1 requires both the factual and counterfactual contact-step lists to remain empty throughout the evaluated future.

S2 instead selects interaction-sensitive cases in which the factual trajectory contains at least one detected contact step and the requested velocity edit produces a nontrivial subsequent trajectory change. The latter condition requires the maximum absolute factual–counterfactual difference over the stored post-anchor primitive states to be at least $10^{-4}$. S2 does not require the counterfactual branch itself to make contact, nor does it require the edit to change contact/non-contact status or contact timing.

Accordingly, S1 and S2 distinguish a contact-free regime from a factual-contact regime in which the edit has a measurable downstream effect; they are evaluated separately rather than treated as complementary exhaustive partitions.

### F.8 Registered thresholds and unit/cell decision rule

The same frozen threshold registry is used by the main low-rank and temporal evaluations. Threshold comparisons are inclusive, and nonfinite unit metrics fail.

Table F4. Registered M1-M5 thresholds for S1 and S2

| Metric | S1 threshold | S2 threshold |
|---|---|---|
| M1 | 0.38876 | 0.46698 |
| M2 | 0.65309 | 0.65681 |
| M3 | 0.28745 | 0.55140 |

| M4 | 0.33938 | 0.25616 |
|---|---|---|
| M5 | 20.26019 | 10.47116 |

For unit $i$ in stratum $s$, $Q_i$ is the joint M1–M5 scientific condition, $E_i$ the numerical/structural eligibility condition, and $J_i$ the final unit-pass condition:

$$Q_i \Leftrightarrow \bigwedge_{k=1}^{5} \left[\text{finite}\left(M_{i,k}\right) \wedge M_{i,k} \leq \theta_{s,k}\right] \qquad E_i \Leftrightarrow M6E_i \wedge M7_i \qquad J_i \Leftrightarrow Q_i \wedge E_i$$

(F17)

$$\text{Coverage} = \frac{N_{\text{joint-pass}}}{N_{\text{expected}}}$$

(F18)

$N_{\text{expected}}$ is 256 for each checkpoint × stratum cell in the main confirmatory assay and 128 for each checkpoint × anchor × stratum cell in the temporal assay. Failed or ineligible units remain in the denominator.

$$\text{CellPass} \Leftrightarrow \text{CountOK} \wedge \text{EligOK} \wedge \text{CoverageOK} \wedge \text{MedianOK} \wedge \text{ControlsPass}$$

(F19)

Here CountOK requires $N_{\text{materialized}} = N_{\text{expected}}$ ; EligOK requires every unit to satisfy $E_i$ ; CoverageOK requires coverage at least 0.8; MedianOK requires every M1–M5 cell median to be finite and within its stratum threshold; and ControlsPass is defined in Appendix G. S1 and S2 are never pooled before this decision.

## F.9 Checkpoint and panel aggregation

A checkpoint passes only if both strata pass:

$$\text{SeedPass}_q = \text{CellPass}_{q,S1} \wedge \text{CellPass}_{q,S2}$$

(F20)

The registered three-checkpoint panel requires at least two passing checkpoints:

$$\text{PanelPass} \Leftrightarrow \left|\left\{q \in \{1,2,3\}: \text{SeedPass}_q\right\}\right| \geq 2$$

(F21)

The temporal tests apply the same rule independently at each anchor or transport target:

$$\begin{aligned} \text{ExistenceOverall} &\Leftrightarrow \bigwedge_{t \in \{5,6,7,8,9\}} \text{PanelPass}_t \\ \text{TransportOverall} &\Leftrightarrow \bigwedge_{t \in \{5,6,8,9\}} \text{PanelPass}_t \end{aligned}$$

(F22)

Trajectory units are not treated as independent model-level replications. Appendix G defines the separate negative-control requirements. The Joint extension does not alter this preregistered Single panel rule.

## F.10 Same-object Joint scoring and specificity reporting

For Joint edits, the two direct velocity errors use the same primitive-coordinate normalization as M1. The Joint direct-target summary is their maximum; M2 and M3 retain their original roles. The unaffected mask has six entries, and the declaration check in M5 permits exactly the two declared velocity changes.

For $a \in \{0,1\}$ indexing horizontal and vertical velocity axes,

$$M_{1,i}^{\text{Joint}} = \max_{a\in\{0,1\}} \frac{\left|\widehat{\mathbf{X}}_{i,0,o_i,2+a} - \mathbf{X}_{i,0,o_i,2+a}^{\text{ref}}\right|}{s_{o_i,2+a}^{P}}$$

(F23)

Native Joint, $\mathbf{U}_4$-oracle, and Addressable Joint routes are compared on the same 101 TEST units per checkpoint. Intended requests are identity-matched to no-patch/sham, random equal-norm, wrong-object, and wrong-vector controls. Appendix H defines the matched aggregation and composition summaries. These later analyses do not modify the preregistered Single thresholds or fresh-replication panel.

## Appendix G. Control Routes and Temporal Generalization Protocols

This appendix defines the reference/control routes and the temporal tests used after the primary Single assay at $t = 7$. Controls distinguish target-specific intervention from generic perturbation, interface or timing errors, and leakage. The temporal tests separate per-anchor recoverability (B1) from direct reuse of the frozen $t = 7$ interface (B2).

### G.1 Positive and structural reference routes

Table G1. Reference and control routes used in the intervention assay

| Route / control | Purpose |
|---|---|
| G0_FACTUAL_NO_EDIT | Checks ordinary factual rollout from the same model and interface. |
| SIMULATOR_COUNTERFACTUAL_TRUTH | External physical reference generated from the edited simulator state. |
| G1_NATIVE_COUNTERFACTUAL_FULL_HIDDEN | Checks whether the model can natively represent and propagate the complete edited current state. |
| G2_FULL_HIDDEN_EQUIVALENCE | Checks that hidden injection reproduces the native counterfactual route to the registered numerical tolerance. |
| SHAM_EDIT | Exercises the edit/map path with zero physical change; should behave as a no-op. |
| RANK_ZERO | Explicit no-patch reference for the low-rank intervention family. |
| RANDOM_ORTHOGONAL_EQUAL_NORM | Tests whether an equal-size but unrelated hidden perturbation can mimic the targeted result. |
| WRONG_OBJECT_EDIT | Tests object specificity by routing the edit to the other object while keeping the edit family comparable. |
| WRONG_TIME | Tests timing specificity by applying the registered mechanism at an adjacent or otherwise registered incorrect anchor. |
| LEAKAGE_NEGATIVE_TEST | Verifies that building the intervention does not require counterfactual future payload. |
| IDENTITY_ORDER_CONTROL | Verifies object-slot identity and decoder/output ordering. |

All reported actions are zero, so action shuffling has no scientific contrast and is marked not applicable. Table G1 supplies the common route vocabulary. In the primary Single assay, registered controls enter rank selection and confirmation. The Joint extension reuses G0-G2 and the structural checks but uses the matched scientific controls in G.5. G1 and G2 are prerequisites, not substitutes for specificity controls.

### G.2 B1: per-anchor carrier existence

B1 refits the same frozen Single rank-4 procedure independently at $t \in \{5, 6, 7, 8, 9\}$. For each checkpoint and anchor, the carrier basis, feature standardization, and ridge-affine map use only that anchor's Single carrier-fit split; evaluation then uses the corresponding sealed cells. B1 therefore tests whether the interface can be recovered at each anchor, not whether numerical carrier coordinates are identical across anchors.

Table G2. Temporal data allocation for the B1 per-anchor analysis

| Temporal data role | Per anchor × stratum | Total across five anchors and two strata |
|---|---|---|
| Carrier fit | 128 units | 1280 units |
| Method development | 128 units | 1280 units |
| Confirmatory nonfinal | 128 units | 1280 units |

## G.3 B2: exact transport of the t = 7 coordinate system

B2 transports, for each checkpoint, the exact $t = 7$ carrier basis, ridge weights and intercept, feature means and scales, rank 4, and $\beta = 1$ to target anchors $t \in \{5, 6, 8, 9\}$. No target-anchor refit, restandardization, rank change, or intervention-strength adjustment is allowed. B1 at the same target is the refitted reference; B2 tests direct reuse of the frozen interface.

Both B1 and B2 use the M1-M5, M6E, M7, control, coverage, and 2-of-3 checkpoint rules in Appendix F. The temporal pairing analysis is descriptive and does not change the cell or panel decision rule.

## G.4 Temporal interpretation boundary

B1 establishes recoverability and B2 direct reuse only over the tested nearby window. Neither implies globally time-invariant latent coordinates, and neither tests temporal transport of the later Joint extension.

## G.5 Matched Joint controls for compositional addressability

Joint addressability is compared with four matched scientific controls: SHAM / NO_PATCH; RANDOM_ORTHOGONAL_EQUAL_NORM; WRONG_OBJECT_JOINT; and MATCHED_WRONG_VECTOR, which keeps the edited object and edit-vector norm fixed while changing direction (by component swap when distinct, otherwise by a one-component sign flip). The last control tests vector-direction specificity beyond edit magnitude.

Intended and control routes are matched by base-unit identity, anchor, edited object, and declared request before scoring on the same Joint units. We report the increase in direct-target and future errors under each control, together with the fraction of matched cases where the control performs worse than the intended route (Appendix H.6). Leakage and identity/order are integrity checks, and wrong-time is secondary. These controls do not affect Single rank selection or the fresh-checkpoint criterion.

## Appendix H. Joint Composition and Addressability Protocols

This appendix defines the matched populations and metrics used for the Joint composition, route-comparison, and specificity analyses in Sections 4.7 and 5.5–5.6. Single edits one velocity component of one object; Joint edits both velocity components of the same object at the same anchor.

### H.1 Exact identity-matched quadruplets and split

For each base unit $i$, the matched routes are factual $F_i$, horizontal Single $S_{x,i}$, vertical Single $S_{y,i}$, and Joint $J_{xy,i}$. They share the same unit identity, edited object, anchor, pre-anchor history, and realized observation noise. The two Single requests sum exactly to the Joint request:

$$\mathbf{e}_{xy,i} = \mathbf{e}_{x,i} + \mathbf{e}_{y,i}. \tag{H1}$$

The declared object and velocity vector are verified from the numerical factual-to-Joint anchor-state difference. The registry contains 512 unique units, split before analysis into 411 FIT and 101 TEST units; row order is never used as an identity key.

Per checkpoint, the full Joint population contains 512 units. The 411 FIT units are used only for factual whitening and local route-operator fitting; the 101 TEST units are used for static/dynamic composition, route comparison, patch geometry, and matched specificity.

### H.2 FIT-only factual-reference whitening

For each checkpoint, whitening is fitted only from factual hidden states in the 411 FIT units. Let $\mathbf{Z}_F$ denote the factual FIT-state matrix. We compute

$$\begin{gathered}\boldsymbol{\mu}_{\mathrm{F}} = \mathrm{mean}(\mathbf{Z}_{\mathrm{F}}) \\ \mathbf{C}_{\mathrm{F}} = \mathrm{cov}(\mathbf{Z}_{\mathrm{F}} - \boldsymbol{\mu}_{\mathrm{F}}) = \mathbf{U}_{\mathrm{wh}}\boldsymbol{\Lambda}_{\mathrm{wh}}\mathbf{U}_{\mathrm{wh}}^{\top} \\ \mathbf{W}_{\mathrm{wh}} = \mathbf{U}_{\mathrm{wh,keep}}\boldsymbol{\Lambda}_{\mathrm{wh,keep}}^{-1/2}\mathbf{U}_{\mathrm{wh,keep}}^{\top}\end{gathered} \tag{H2}$$

Here, $\boldsymbol{\mu}_{\mathrm{F}}$ and $\mathbf{C}_{\mathrm{F}}$ are the mean and covariance of the factual FIT hidden states. $\mathbf{U}_{\mathrm{wh}}$ contains the covariance eigenvectors and $\boldsymbol{\Lambda}_{\mathrm{wh}} = diag(\lambda_{wh,1}, \lambda_{wh,2}, \ldots)$ the corresponding eigenvalues. An eigen-direction $j$ is retained when $\lambda_{\mathrm{wh,j}}/\lambda_{\mathrm{wh,max}} > 10^{-10}$, where $\lambda_{\mathrm{wh,max}} = max_j \lambda_{wh,j}$. The subscript “keep” denotes the retained eigenvectors and their corresponding eigenvalues. Treating hidden states as column vectors, the frozen TEST-time map is

$$\mathrm{wh}(\mathbf{z}) = \mathbf{W}_{\mathrm{wh}}(\mathbf{z} - \boldsymbol{\mu}_{\mathrm{F}}) \tag{H3}$$

This whitening only defines a checkpoint-specific factual reference gauge. It is not fitted on TEST units and does not construct or refit the original $\mathbf{U}_4$ carrier or the addressable intervention.

### H.3 Static Joint composition

For TEST unit $i$, define the whitened responses relative to the same factual baseline:

$$\begin{gathered}\Delta\mathbf{z}_{x,i}^{\mathrm{wh}} = \mathrm{wh}(\mathbf{z}_{S_x,i}) - \mathrm{wh}(\mathbf{z}_{F,i}) \\ \Delta\mathbf{z}_{y,i}^{\mathrm{wh}} = \mathrm{wh}\left(\mathbf{z}_{S_y,i}\right) - \mathrm{wh}(\mathbf{z}_{F,i})\end{gathered}$$

$$\Delta\mathbf{z}_{xy,i}^{\mathrm{wh}} = \mathrm{wh}\left(\mathbf{z}_{J_{xy},i}\right) - \mathrm{wh}\left(\mathbf{z}_{F,i}\right)$$
$$\Delta\mathbf{z}_{\mathrm{add},i}^{\mathrm{wh}} = \Delta\mathbf{z}_{x,i}^{\mathrm{wh}} + \Delta\mathbf{z}_{y,i}^{\mathrm{wh}}$$

(H4)

Here $x$ and $y$ denote horizontal and vertical velocity edits, not position edits. The primary static composition error is

$$E_{\mathrm{add},i} = \frac{2\left\|\Delta\mathbf{z}_{xy,i}^{\mathrm{wh}} - \Delta\mathbf{z}_{\mathrm{add},i}^{\mathrm{wh}}\right\|_2}{\left\|\Delta\mathbf{z}_{xy,i}^{\mathrm{wh}}\right\|_2 + \left\|\Delta\mathbf{z}_{\mathrm{add},i}^{\mathrm{wh}}\right\|_2 + \varepsilon}, \qquad \varepsilon = 10^{-12}$$

(H5)

Lower $E_{\mathrm{add},i}$ indicates a more additive native Joint response. Supporting magnitude and direction diagnostics are

$$R_{\mathrm{norm},i} = \frac{\left\|\Delta\mathbf{z}_{xy,i}^{\mathrm{wh}}\right\|_2}{\left\|\Delta\mathbf{z}_{\mathrm{add},i}^{\mathrm{wh}}\right\|_2 + \varepsilon}$$
$$r_{\mathrm{norm},i} = \left|\log\left(\max\left(R_{\mathrm{norm},i}, \varepsilon\right)\right)\right|$$
$$\cos_{\mathrm{add},i} = \frac{\left(\Delta\mathbf{z}_{xy,i}^{\mathrm{wh}}\right)^{\top}\Delta\mathbf{z}_{\mathrm{add},i}^{\mathrm{wh}}}{\left\|\Delta\mathbf{z}_{xy,i}^{\mathrm{wh}}\right\|_2\left\|\Delta\mathbf{z}_{\mathrm{add},i}^{\mathrm{wh}}\right\|_2 + \varepsilon}$$
$$\cos_{xy,i} = \frac{\left(\Delta\mathbf{z}_{x,i}^{\mathrm{wh}}\right)^{\top}\Delta\mathbf{z}_{y,i}^{\mathrm{wh}}}{\left\|\Delta\mathbf{z}_{x,i}^{\mathrm{wh}}\right\|_2\left\|\Delta\mathbf{z}_{y,i}^{\mathrm{wh}}\right\|_2 + \varepsilon}$$

(H6)

$E_{\mathrm{add}}$ is the primary static quantity; the norm and cosine descriptors are supporting diagnostics only.

### H.4 Dynamic composition under the actual recurrence

For TEST unit $i$, form the synthetic additive initialization and the native Joint initialization:

$$\mathbf{z}_{\mathrm{add},i}^{(0)} = \mathbf{z}_{S_x,i} + \mathbf{z}_{S_y,i} - \mathbf{z}_{F,i}$$
$$\mathbf{z}_{\mathrm{joint},i}^{(0)} = \mathbf{z}_{J_{xy},i}$$

(H7)

Let $T$ denote one exact autonomous recurrent transition of the frozen checkpoint with the recorded zero action, no future observation assimilation, no state clamp, and no hidden intervention. With $T^0$ the identity and $k \in \{0,1,3,6,12\}$,

$$\mathbf{z}_{\mathrm{add},i}^{(k)} = T^k\left(\mathbf{z}_{\mathrm{add},i}^{(0)}\right), \qquad \mathbf{z}_{\mathrm{joint},i}^{(k)} = T^k\left(\mathbf{z}_{\mathrm{joint},i}^{(0)}\right)$$
$$\mathbf{z}_{\mathrm{add},i}^{\mathrm{wh},(k)} = \mathrm{wh}\left(\mathbf{z}_{\mathrm{add},i}^{(k)}\right), \qquad \mathbf{z}_{\mathrm{joint},i}^{\mathrm{wh},(k)} = \mathrm{wh}\left(\mathbf{z}_{\mathrm{joint},i}^{(k)}\right)$$
$$E_{\mathrm{dyn},i}(k) = \frac{2\left\|\mathbf{z}_{\mathrm{add},i}^{\mathrm{wh},(k)} - \mathbf{z}_{\mathrm{joint},i}^{\mathrm{wh},(k)}\right\|_2}{\left\|\mathbf{z}_{\mathrm{add},i}^{\mathrm{wh},(k)}\right\|_2 + \left\|\mathbf{z}_{\mathrm{joint},i}^{\mathrm{wh},(k)}\right\|_2 + \varepsilon}$$

(H8)

The four $\mathbf{z}$ terms in Eq. (H7) are the 192-D native post-anchor states for the factual, matched Single-$x$, matched Single-$y$, and Joint routes. The additive state is a synthetic diagnostic initialization and need not lie on a naturally visited hidden-state manifold. Lower $E_{\mathrm{dyn},i}(k)$ means the additive initialization remains closer to the native Joint state after $k$ recurrent transitions. This diagnostic does not use $\mathbf{U}_4$, does not roll out the fitted operators in Appendix I, and does not establish an exact linear composition law.

### H.5 Same-101 Native Joint, U4-oracle, and Addressable Joint comparison

For each MF1 and MF2 checkpoint, Native Joint, privileged $\mathbf{U}_4$-oracle, and Addressable Joint are evaluated on the same 101 TEST units. For route $r$ and unit $i$, define

$$I_{r,i}^{\mathrm{pass}} = \mathbb{1}[\mathrm{JointPass}(r,i)],$$

$$C_r = \frac{1}{101}\sum_{i=1}^{101} I_{r,i}^{\mathrm{pass}},$$

$$e_{r,i,t} = \mathrm{RMSE}\big(\hat{\mathbf{y}}_{r,i,t}, \mathbf{y}_{i,t}\big),$$

$$R_{r,t} = \mathrm{median}_i\, e_{r,i,t}.$$

(H9)

$C_r$ is route coverage and $R_{r,t}$ is the median primitive-state rollout error. Checkpoint summaries are computed first; family summaries are then taken across the three checkpoints. Across MF1 and MF2, the common TEST comparison contains $6 \times 101 = 606$ unique checkpoint-unit pairs with no duplicates. The pass indicator $I_{r,i}^{\mathrm{pass}}$ is used here to avoid confusion with the carrier projector $\mathbf{P}$ defined in Appendix I.

### H.6 Matched Joint specificity

For metric $m$ (Joint M1, M3, or terminal RMSE), compare the intended addressable route with each matched control on the same TEST units:

$$\Delta_m = \mathrm{median}_i m_{c,i} - \mathrm{median}_i m_{a,i},$$

$$w_m = \frac{1}{101}\sum_{i=1}^{101} \mathbb{1}\big[m_{c,i} > m_{a,i}\big].$$

(H10)

Here $m_{c,i}$ and $m_{a,i}$ are the control and intended-route errors, $\Delta_m > 0$ means the control has larger median error, and $w_m$ is the paired-worse fraction. Controls are no-patch/sham, random orthogonal equal-norm, wrong-object Joint, and matched wrong-vector. Unit-level pairs support within-checkpoint specificity; checkpoints remain the replication level.

### H.7 Interpretation boundary

Static $E_{\mathrm{add}}$ is the primary composition diagnostic; dynamic composition is supporting evidence. MF0/MF1/MF2 comparisons are descriptive family-level associations under matched budgets, not universal causal claims about training.

## Appendix I. Carrier-Relative and Recurrent-Dynamics Diagnostics

This appendix defines post-hoc diagnostics used to distinguish a compact intervention-entry interface from a closed four-dimensional dynamical state. All analyses are checkpoint-specific, use the FIT-only factual-reference whitening from Appendix H.2, and do not alter the original rank selection or confirmatory decisions.

### I.1 Whitened $\mathbf{U}_4$ projector and complement

Let $\mathbf{U}_4 \in \mathbb{R}^{192\times 4}$ be the checkpoint-specific Single-derived carrier in raw hidden coordinates, and let $\mathbf{W}_{\mathrm{wh}}$ be the FIT-only whitening matrix from Eq. (H2). We first orthonormalize the whitened carrier and then define its projector and complement:

$$\mathbf{Q}_U = \mathrm{orth}(\mathbf{W}_{\mathrm{wh}}\mathbf{U}_4)$$
$$\mathbf{P} = \mathbf{Q}_U\mathbf{Q}_U^{\top}$$
$$\mathbf{Q} = \mathbf{I}_{192} - \mathbf{P}$$

(I1)

Here $\mathbf{P}$ projects onto the four-dimensional carrier in factual-whitened coordinates and $\mathbf{Q}$ onto its 188-dimensional orthogonal complement. These projectors are diagnostic only and do not refit $\mathbf{U}_4$.

### I.2 Composition inside $\mathbf{U}_4$ and its complement

For TEST unit $i$, let $\Delta\mathbf{z}_{xy,i}^{\mathrm{wh}}$ and $\Delta\mathbf{z}_{\mathrm{add},i}^{\mathrm{wh}}$ be the native Joint and additive Single-derived responses from Appendix H.3, and define the composition defect

$$\mathbf{d}_i = \Delta\mathbf{z}_{xy,i}^{\mathrm{wh}} - \Delta\mathbf{z}_{\mathrm{add},i}^{\mathrm{wh}}$$

The carrier and complement components are

$$\mathbf{a}_{U,i} = \mathbf{P}\Delta\mathbf{z}_{xy,i}^{\mathrm{wh}} \qquad \mathbf{b}_{U,i} = \mathbf{P}\Delta\mathbf{z}_{\mathrm{add},i}^{\mathrm{wh}} \qquad \mathbf{d}_{U,i} = \mathbf{P}\mathbf{d}_i$$
$$\mathbf{a}_{Q,i} = \mathbf{Q}\Delta\mathbf{z}_{xy,i}^{\mathrm{wh}} \qquad \mathbf{b}_{Q,i} = \mathbf{Q}\Delta\mathbf{z}_{\mathrm{add},i}^{\mathrm{wh}} \qquad \mathbf{d}_{Q,i} = \mathbf{Q}\mathbf{d}_i$$
$$E_{\mathrm{add},U,i} = \frac{2\,\|\,\mathbf{d}_{U,i}\,\|_2}{\|\,\mathbf{a}_{U,i}\,\|_2 + \|\,\mathbf{b}_{U,i}\,\|_2 + \varepsilon}$$
$$E_{\mathrm{add},Q,i} = \frac{2\,\|\,\mathbf{d}_{Q,i}\,\|_2}{\|\,\mathbf{a}_{Q,i}\,\|_2 + \|\,\mathbf{b}_{Q,i}\,\|_2 + \varepsilon}$$

(I2)

For any whitened response or defect vector $\mathbf{v}$, define the squared-norm fractions and dimension-normalized enrichments

$$f_U(\mathbf{v}) = \frac{\|\,\mathbf{P}\mathbf{v}\,\|_2^2}{\|\,\mathbf{v}\,\|_2^2 + \varepsilon} \qquad f_Q(\mathbf{v}) = \frac{\|\,\mathbf{Q}\mathbf{v}\,\|_2^2}{\|\,\mathbf{v}\,\|_2^2 + \varepsilon}$$
$$\mathrm{enrich}_U(\mathbf{v}) = \frac{f_U(\mathbf{v})}{4/192} \qquad \mathrm{enrich}_Q(\mathbf{v}) = \frac{f_Q(\mathbf{v})}{188/192}$$

(I3)

Here, enrichment measures concentration relative to the dimensional baseline; it is not a significance test and does not imply that $\mathbf{U}_4$ is the unique locus of composition.

## I.3 Local route-conditioned recurrent operators

For route $r \in \{F, S, J\}$ (Factual, Single, Joint), consecutive factual-whitened hidden states from the 411 FIT units are stacked as row matrices $\mathbf{X}_r$ and $\mathbf{Y}_r$. With an affine intercept,

$$\mathbf{Z}_r = [\mathbf{X}_r, \mathbf{1}]$$
$$\mathbf{B}_r = \mathrm{pinv}(\mathbf{Z}_r; \mathrm{rcond} = 10^{-12})\mathbf{Y}_r$$
$$\begin{bmatrix} \mathbf{K}_r^\top \\ \mathbf{c}_r^\top \end{bmatrix} = \mathbf{B}_r$$
$$\widehat{\mathbf{Y}}_r = \mathbf{X}_r \mathbf{K}_r^\top + \mathbf{1}\mathbf{c}_r^\top$$
$$R_r^2 = 1 - \frac{\mathrm{SSE}_r}{\mathrm{SST}_r} \tag{I4}$$

where $\mathrm{SSE}_r = \| \mathbf{Y}_r - \widehat{\mathbf{Y}}_r \|_F^2$ and $\mathrm{SST}_r$ is the corresponding total sum of squares with the registered numerical floor. The fit is evaluated on the 101 TEST units.

The Single-to-Joint operator shift and supporting descriptors are

$$\Delta\mathbf{K}_{JS} = \mathbf{K}_J - \mathbf{K}_S$$
$$d_{JS} = \frac{\| \mathbf{K}_J - \mathbf{K}_S \|_F}{\| \mathbf{K}_F \|_F + \varepsilon}$$
$$\rho(\mathbf{K}) = \max_i |\lambda_i(\mathbf{K})|$$
$$\nu(\mathbf{K}) = \frac{\| \mathbf{K}^\top\mathbf{K} - \mathbf{K}\mathbf{K}^\top \|_F}{\| \mathbf{K} \|_F^2 + \varepsilon} \tag{I5}$$

Here $\|\cdot\|_F$ denotes the Frobenius norm. These operators are local descriptive fits over the registered population, not global Koopman operators, exact recurrent dynamics, or reduced-state models.

## I.4 P/Q blocks, retention, leakage, and inward transfer

For any operator $\mathbf{A}$, including $\Delta\mathbf{K}_{JS}$, define

$$\mathbf{A}_{PP} = \mathbf{PAP} \qquad \mathbf{A}_{PQ} = \mathbf{PAQ}$$
$$\mathbf{A}_{QP} = \mathbf{QAP} \qquad \mathbf{A}_{QQ} = \mathbf{QAQ} \tag{I6}$$

For a route operator $\mathbf{K}$, carrier retention, outward leakage, and inward transfer are

$$R_{\mathrm{in}} = \frac{\| \mathbf{PKP} \|_F^2}{\| \mathbf{KP} \|_F^2 + \varepsilon}$$
$$L_{\mathrm{out}} = \frac{\| \mathbf{QKP} \|_F^2}{\| \mathbf{KP} \|_F^2 + \varepsilon}$$
$$G_{\mathrm{in}} = \frac{\| \mathbf{PKQ} \|_F^2}{\| \mathbf{KQ} \|_F^2 + \varepsilon} \tag{I7}$$

When the denominator is nonzero, $R_{\text{in}} + L_{\text{out}} \approx 1$ up to numerical precision. Low retention and high leakage indicate that the fitted one-step response to a carrier-originating perturbation is dominated by coupling into the wider hidden state rather than remaining inside $\mathbf{U}_4$.

### I.5 Dimension-normalized operator blocks

Because raw Frobenius magnitude scales with block size, for block $b \in \{PP, PQ, QP, QQ\}$ with $n_b$ entries we report

$$f_b = \frac{\| \mathbf{A}_b \|_F^2}{\| \mathbf{A} \|_F^2 + \varepsilon}$$

$$\text{RMS}_b = \frac{\| \mathbf{A}_b \|_F}{\sqrt{n_b}} \qquad \text{RMS}_{\text{global}} = \frac{\| \mathbf{A} \|_F}{192}$$

$$\text{RMSenrich}_b = \frac{\text{RMS}_b}{\text{RMS}_{\text{global}} + \varepsilon}$$

$$\text{EnergyEnrich}_b = \frac{f_b}{n_b/192^2}$$

(I8)

The block sizes are $n_{PP} = 16$, $n_{PQ} = n_{QP} = 752$, and $n_{QQ} = 35{,}344$. Here, "energy" denotes squared Frobenius magnitude of the fitted operator block, not physical energy. These descriptors compare per-entry magnitude or concentration after accounting for block dimension; they are not hypothesis tests.

### I.6 Full, baseline, and request-only Joint patch geometry

For the rank-4 Single-only addressable map $f_{\text{addr},4}$, factual state $\mathbf{s}^F$, and Joint request $\mathbf{e}_{xy}$,

$$\mathbf{c}_{\text{full}} = f_{\text{addr},4}(\mathbf{s}^F, \mathbf{e}_{xy})$$

$$\mathbf{c}_{\text{base}} = f_{\text{addr},4}(\mathbf{s}^F, \mathbf{0})$$

$$\mathbf{c}_{\text{request}} = \mathbf{c}_{\text{full}} - \mathbf{c}_{\text{base}}$$

$$\delta\mathbf{z}_{\text{full}} = \mathbf{U}_4\mathbf{c}_{\text{full}} \qquad \delta\mathbf{z}_{\text{base}} = \mathbf{U}_4\mathbf{c}_{\text{base}} \qquad \delta\mathbf{z}_{\text{request}} = \mathbf{U}_4\mathbf{c}_{\text{request}}$$

(I9)

Thus $\delta\mathbf{z}_{\text{full}} = \delta\mathbf{z}_{\text{base}} + \delta\mathbf{z}_{\text{request}}$. The actual rollout uses $\delta\mathbf{z}_{\text{full}}$; the other two components are geometry diagnostics only.

We treat the full affine patch as the primary geometry object because it is the intervention actually used for rollout. The baseline and request-only components are secondary post-hoc decompositions used only to interpret the affine patch.

Let $\delta\mathbf{z}_{\text{oracle}}$ be the projected native Joint hidden difference from Eq. (D5a). For any raw-space perturbation $\mathbf{v}$, define its four-dimensional coordinate in the whitened carrier by

$$\mathbf{u}(\mathbf{v}) = \mathbf{Q}_U^\top \mathbf{W}_{\text{wh}}\mathbf{v}$$

$$E(\mathbf{v}, \text{oracle}) = \frac{2 \| \mathbf{u}(\mathbf{v}) - \mathbf{u}(\delta\mathbf{z}_{\text{oracle}}) \|_2}{\| \mathbf{u}(\mathbf{v}) \|_2 + \| \mathbf{u}(\delta\mathbf{z}_{\text{oracle}}) \|_2 + \varepsilon}$$

(I10)

The comparison is computed for $\mathbf{v} \in \{\delta\mathbf{z}_{\text{full}}, \delta\mathbf{z}_{\text{base}}, \delta\mathbf{z}_{\text{request}}\}$ on the same 101 TEST units. Distance to this particular projected native Joint oracle is a bounded geometry diagnostic, not a sufficient proxy for rollout effectiveness.

### I.7 Interpretation boundary

These diagnostics support only bounded mechanistic interpretations. Carrier enrichment does not make $\mathbf{U}_4$ the unique locus of composition; local operator coupling does not establish a globally invariant four-dimensional state; and oracle-distance results do not imply that hidden-space geometry is irrelevant. All conclusions remain checkpoint-, population-, horizon-, and diagnostic-specific.